\documentclass[10pt,twocolumn,letterpaper]{article}

\usepackage[pagenumbers]{cvpr} 

\definecolor{cvprblue}{rgb}{0.21,0.49,0.74}
\usepackage[pagebackref,breaklinks,colorlinks,allcolors=cvprblue]{hyperref}

\usepackage{multirow}
\usepackage{graphicx}
\usepackage{subcaption}
\usepackage{algorithm}               
\usepackage{algpseudocode}
\usepackage{setspace}
\usepackage{enumitem}
\usepackage{color, colortbl}
\usepackage{amsmath}
\usepackage{amssymb}
\usepackage{mathtools}
\usepackage{natbib}

\title{Learning to Sample Effective and Diverse Prompts for Text-to-Image Generation}

\author{
{Taeyoung Yun$^{1}$\thanks{Work done during TY's visit to HKUST.} ~ Dinghuai Zhang$^{2}$ ~ Jinkyoo Park$^{1}$ ~ Ling Pan$^{3}$} 
\\
{$^{1}$ Korea Advanced Institute of Science and Technology $^{2}$ Microsoft Research} \\ 
$^{3}$ Hong Kong University of Science and Technology
}

\begin{document}
\maketitle

\begin{abstract}
Recent advances in text-to-image diffusion models have achieved impressive image generation capabilities. 
However, it remains challenging to control the generation process with desired properties (e.g., aesthetic quality, user intention), which can be expressed as black-box reward functions. 
In this paper, we focus on prompt adaptation, which refines the original prompt into model-preferred prompts to generate desired images. While prior work uses reinforcement learning (RL) to optimize prompts, we observe that applying RL often results in generating similar postfixes and deterministic behaviors.
To this end, we introduce \textbf{P}rompt \textbf{A}daptation with \textbf{G}FlowNets (\textbf{PAG}), a novel approach that frames prompt adaptation as a probabilistic inference problem. 
Our key insight is that leveraging Generative Flow Networks (GFlowNets) allows us to shift from reward maximization to sampling from an unnormalized density function, enabling both high-quality and diverse prompt generation.
However, we identify that a naive application of GFlowNets suffers from mode collapse and uncovers a previously overlooked phenomenon: the progressive loss of neural plasticity in the model, which is compounded by inefficient credit assignment in sequential prompt generation. To address this critical challenge, we develop a systematic approach in PAG with flow reactivation, reward-prioritized sampling, and reward decomposition for prompt adaptation.
Extensive experiments validate that PAG successfully learns to sample effective and diverse prompts for text-to-image generation. 
We also show that PAG exhibits strong robustness across various reward functions and transferability to different text-to-image models. 
\end{abstract}

\section{Introduction}
\begin{figure}[t]
    \centering
    \includegraphics[width=0.95\linewidth]{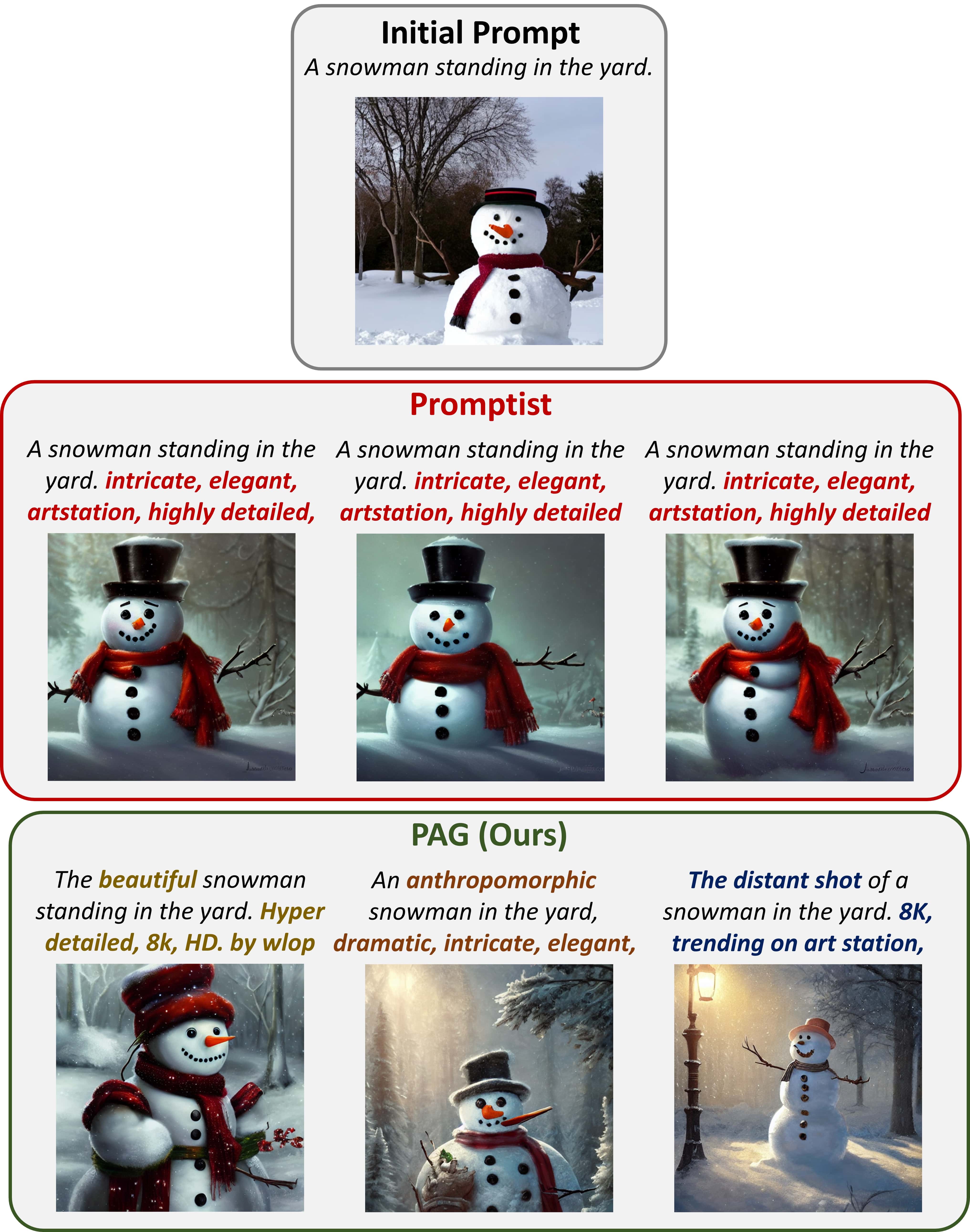}
    \caption{Comparison of adapted prompts and their corresponding images of Prompist~\cite{hao2024optimizing} (based on reward-maximizing RL) and our method, PAG. While Promptist leads to mode collapse in the prompt space and converges to similar outputs, PAG achieves high image quality while painting generation diversity.}
    \label{fig:main_figure_preview}
    \vspace{-18pt}
\end{figure}

Recent advances in diffusion models \citep{sohl2015deep, ho2020denoising}, combined with pre-trained text encoders \citep{raffel2020exploring, radford2021learning} have shown remarkable capability in generating creative and photorealistic images conditioned on novel prompts \citep{rombach2022high, ramesh2022hierarchical, saharia2022photorealistic}. However, generating images with desired properties (e.g., aesthetic quality, user intention) remains challenging, as these models are typically optimized for likelihood maximization over training data distributions~\citep{ho2020denoising}.

To further improve the generation process with desired properties, there has been a line of work focusing on fine-tuning diffusion models with human feedback~\citep{blacktraining, fan2024reinforcement,prabhudesai2023aligning, clarkdirectly, zhang2024improving}.
While these methods demonstrate promising results, they rely on access to model parameters, rendering them incompatible with several state-of-the-art closed-source models~\citep{ramesh2022hierarchical, saharia2022photorealistic}.
Moreover, as text-to-image diffusion models continue to grow exponentially in size, the computational cost of direct fine-tuning becomes prohibitively expensive. This challenge is further compounded by the need for model-specific retraining across different text-to-image diffusion models.

In contrast, prompt adaptation has emerged as a promising alternative~\citep{hao2024optimizing, kim2023multiprompter, wang2024discrete}, which aims to improve the initial prompt for generating images with desired properties. This approach eliminates the need for access to model parameters, enabling zero-shot transfer to different text-to-image diffusion models.
Notably, Promptist \citep{hao2024optimizing} employs reinforcement learning (RL) to fine-tune language models for prompt adaptation. While it shows promising results, our analysis reveals a critical limitation: its reward-maximizing principle tends to concentrate the policy on a narrow, high-reward region. This often leads to a deterministic policy that generates similar postfixes, as demonstrated in Figure~\ref{fig:main_figure_preview}. 
Such deterministic behavior can reduce the method to simple heuristics, significantly hindering its generalization capacity across different prompt types and text-to-image diffusion models. It underscores the critical need for approaches that can generate both effective and diverse adapted prompts for text-to-image generation.

In this paper, we introduce \textbf{P}rompt \textbf{A}daptation with \textbf{G}FlowNets (\textbf{PAG}), a novel approach that addresses these fundamental challenges by reformulating prompt adaptation as a probabilistic inference~\citep{bengio2021flow}. Our approach leverages Generative Flow Networks (GFlowNets, ~\citep{bengio2023gflownet}) to learn a generative policy that samples from an unnormalized reward distribution~\citep{bengio2021flow}, which is well-suited for this scenario.
While this approach shows initial promise, we observe that naively fine-tuning language models with GFlowNets for prompt adaptation suffers from a mode collapse issue. Our investigation uncovers a previously unrecognized phenomenon in GFlowNet fine-tuning that mirrors a key principle in neuroscience: gradual hardening of brain circuits~\citep{livingston1966brain, mateos2019impact}. In other words, GFlowNets agent experiences a progressive loss of neural plasticity as certain neural pathways become inactive, diminishing its capacity to learn from and adapt to diverse patterns. Furthermore, inefficient credit assignment across sequential prompt generation process leads to insufficient sample efficiency and ultimately exacerbates mode collapse issue.

To this end, we introduce our novel components in PAG to systematically address these critical challenges in prompt adaptation for text-to-image modeling. 
First, we propose a flow reactivation mechanism to revive dormant neural pathways in the GFlowNets agent, complemented by reward-prioritized sampling to effectively consolidate of high-quality experiences. These two components jointly achieve adaptation flexibility while maintaining training stability.
Building upon this foundation, we develop a progressive reward decomposition scheme in our framework to provide fine-grained learning signals at intermediate generation steps, which enables more precise credit assignment throughout the training procedure.

Our contributions can be summarized as below:
\begin{itemize}
    \item We introduce PAG, a novel framework that reformulates prompt adaptation as a probabilistic inference problem, leveraging GFlowNets to generate both effective and diverse adapted prompts.
    \item We identify and address a previously overlooked challenge in GFlowNets fine-tuning - the progressive loss of neural plasticity leading to mode collapse - through a systematic approach that maintains network expressivity and ensures precise credit assignment.
    \item Extensive experiments show that PAG generates both effective and diverse adapted prompts, exhibits robustness to different reward functions, and enables effective zero-shot transfer to various text-to-image diffusion models. 
\end{itemize}

\section{Preliminaries}
\subsection{Prompt Adaptation}
Let $p_{\theta}$ denote a pre-trained language model that generates improved prompt $\mathbf{y}$ conditioned on the initial prompt $\mathbf{x}$, i.e., $\mathbf{y}\sim p_{\theta}(\cdot\vert\mathbf{x})$. A text-to-image diffusion model $p_{\psi}$ then generates images given text inputs $\mathbf{y}$. Following~\citep{hao2024optimizing}, a reward function $r(\mathbf{x}, \mathbf{y})$ for prompt adaptation is defined as follows:
\begin{align}\label{eq:task_reward}
    &\mathbb{E}_{i_\mathbf{x}\sim p_{\psi}(\cdot\vert\mathbf{x}), i_\mathbf{y}\sim p_{\psi}(\cdot\vert\mathbf{y})}\left[r_{\text{aes}}(i_\mathbf{x}, i_\mathbf{y}) + r_{\text{rel}}(\mathbf{x}, i_\mathbf{y})\right], 
\end{align}
where $r_{\text{aes}}(i_{\mathbf{x}}, i_{\mathbf{y}})=g_{\text{aes}}(i_\mathbf{y}) - g_{\text{aes}}(i_\mathbf{x})$ measures the improvement in aesthetic quality of images using the LAION aesthetic predictor \citep{laion}, $g_{\text{aes}}$. The term $r_{\text{rel}}(\mathbf{x}, i_{\mathbf{y}})=\min(20.0\times g_{\text{CLIP}}(\mathbf{x}, i_{\mathbf{y}}) - 5.6, 0.0)$ quantifies the relevance between generate images from $\mathbf{y}$ and the initial prompt $\mathbf{x}$ using the CLIP similarity function, $g_{\text{CLIP}}$ \citep{radford2021learning}. Our objective is fine-tuning language model parameters $\theta$ to optimize the target reward function:
\begin{align}
\label{eq:objective}
\mathbb{E}_{\mathbf{x}\sim\mathcal{D}}\left[\mathbb{E}_{\mathbf{y}\sim p_{\theta}(\cdot\vert\mathbf{x})}\left[r(\mathbf{x}, \mathbf{y})\right]-\beta\cdot D_{\text{KL}}(p_{\theta}(\cdot\vert\mathbf{x})\Vert p_{\text{ref}}(\cdot\vert\mathbf{x}))\right],
\end{align}
where the $D_{\text{KL}}$ term enforces the generated prompts to be close to natural language that human can understand.

\subsection{Generative Flow Networks (GFlowNets)} \label{sec:bg}
GFlowNets are a family of probabilistic methods that sample compositional objects proportionally to an unnormalized distribution defined by a reward function~\citep{bengio2021flow,bengio2023gflownet}. 
Let $\mathcal{S}$ denote the state space and $\mathcal{A}$ the action space, forming nodes and edges in a directed acyclic graph. We define a unique initial state $s_0$ without incoming edges and set of terminal states $\mathcal{X}\subset\mathcal{S}$ without outgoing edges. A sequence from the initial state $s_0$ to the terminal state $x\in\mathcal{X}$ is called a trajectory, denoted as $\tau=(s_0\rightarrow s_1\rightarrow \cdots\rightarrow s_T=x)$, which is generated sequentially.
Let $R:\mathcal{X}\rightarrow\mathbb{R}_{\geq0}$ be a non-negative reward function defined on terminal states. GFlowNets aim to train a stochastic policy $P_{F}$ that generates samples proportional to the reward, i.e., $P_{F}^{T}(x)\propto R(x)$, where $P_{F}^{T}(x)=\sum_{\tau_{\rightarrow x}}P_{F}(\tau)$ is the marginal likelihood of sampling trajectories that result in $x$ ($\tau_{\rightarrow x}$) from the forward policy:
\begin{align}
    P_F^{T}(x)=\sum_{\tau_{\rightarrow x}}\prod_{t=1}^{T}P_{F}(s_t\vert s_{t-1})\propto R(x).
\end{align}
In practice, GFlowNets can be trained by parameterizing the forward policy with a neural network, $P_{F}(s_t\vert s_{t-1};\theta)$, using various training objectives.

\vspace{5pt}
\noindent \textbf{Trajectory Balance (TB, \citep{malkin2022trajectory})} TB introduces an additional backward policy $P_{B}(s_{t-1}\vert s_t;\theta)$ that models the distribution of parents given a child state and a total flow $Z_{\theta}$ to approximate the partition function. Given a trajectory $\tau=(s_0\rightarrow\cdots\rightarrow s_T=x)$, TB aims to minimize the loss in Eq.~(\ref{eq:tb_loss}). If the loss becomes zero for all possible trajectories, it implies that $P_{F}^{T}(x)\propto R(x)$.
\begin{align}
    \mathcal{L}(\tau;\theta)=\left(\log\frac{Z_{\theta}\prod_{t=1}^{T}P_{F}(s_{t}\vert s_{t-1};\theta)}{R(x)\prod_{t=1}^{T}P_{B}(s_{t-1}\vert s_{t};\theta)}\right)^2.
    \label{eq:tb_loss}
\end{align}
\noindent \textbf{Detailed Balance (DB, \citep{bengio2023gflownet})} DB considers flow matching at the edge level instead of the trajectory level and introduces state flow function $F_{\theta}:\mathcal{S}\rightarrow\mathbb{R}_{\geq0}$, which approximates the total flow through state $s$. 
Given an intermediate transition $(s_{t-1}\rightarrow s_t)$, DB aims to minimize the loss in Eq.~(\ref{eq:db_loss}), with $F_{\theta}(s_t)$ replaced by the terminal reward $R(x)$ at terminal states for $t=T$ (i.e., $F_{\theta}(x)=R(x)$).
\begin{align}
    &\mathcal{L}(s_{t-1}, s_{t};\theta)=\left(\log\frac{F_{\theta}(s_{t-1})P_{F}(s_{t}\vert s_{t-1};\theta)}{F_{\theta}(s_{t})P_{B}(s_{t-1}\vert s_{t};\theta)}\right)^2.
    \label{eq:db_loss}
\end{align}
It is often challenging to train GFlowNets due to delayed credit assignment from terminal-only reward signals \citep{pan2023better, jang2024learning}. The forward-looking (FL) technique addresses this by extending rewards to all states and optimizing $\tilde{F_{\theta}}(s)$ in the sense that $F_{\theta}(s)=\tilde{F_{\theta}}(s)R(s)$ for all states, which is a reparameterization of flows in DB.
The resulting FL-DB objective is to minimize the following loss:
\begin{align}
\label{eq:fl-db-objective}
    &\mathcal{L}(s_{t-1}, s_{t};\theta)=\left(\log\frac{\tilde{F_{\theta}}(s_{t-1})P_{F}(s_{t}\vert s_{t-1};\theta)R(s_{t-1})}{\tilde{F_{\theta}}(s_{t})P_{B}(s_{t-1}\vert s_{t};\theta)R(s_t)}\right)^2.
\end{align}

\subsection{Dormant Neuron Phenomenon}
Recent studies have revealed that scaling deep RL networks faces challenges due to parameter under-utilization~\citep{kumarimplicit, lyleunderstanding, sokar2023dormant}. Following \citet{sokar2023dormant}, we quantify neural plasticity by tracking dormant neurons during the training progresses defined as below, where a neuron $i$ in layer $\ell$ is {${\tau}$-dormant} if its activation score is $s^{\ell}_{i}\leq\tau$. 

\vspace{5pt}
\noindent \textbf{Definition 2.3.1.} Given input dataset $\mathcal{D}$, let $h_i(\mathbf{x})$ be the activation of neuron $i$ in layer $\ell$ under input $\mathbf{x}\in\mathcal{D}$, then its activation score is defined as follows:
\begin{align}\label{eq:dormant}
    s^{\ell}_{i}=\frac{\mathbb{E}_{\mathbf{x}\in\mathcal{D}}\vert h^{\ell}_{i}(\mathbf{x})\vert}{\frac{1}{H^{\ell}}\sum_{k\in h}\mathbb{E}_{\mathbf{x}\in\mathcal{D}}\vert h^{\ell}_{k}(\mathbf{x})\vert}.
\end{align}
where $H^l$ is the number of hidden units in the $l$th layer.

\section{Prompt Adaptation with GFlowNets}
\begin{figure*}[t]
    \centering
    \includegraphics[width=0.9\textwidth]{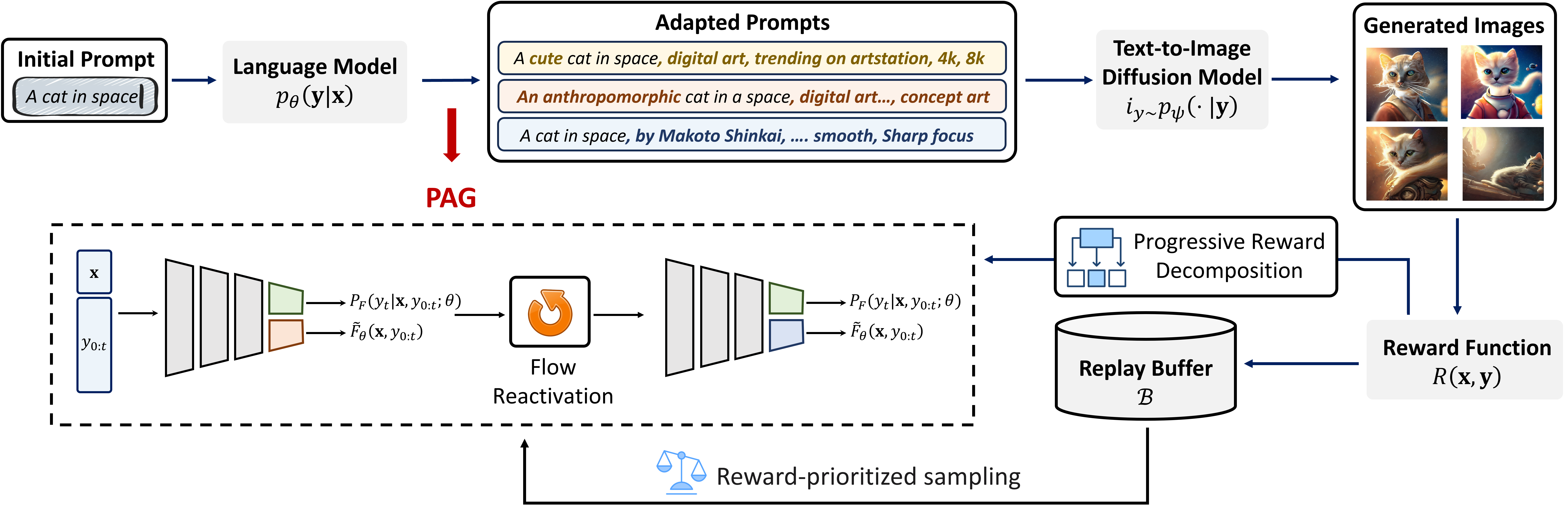}
    \caption{The high-level illustration of PAG. Given an initial prompt, LM generates adapted prompts by PAG. Then, we generate images from prompts and get a reward. Using observations, we fine-tune LM as a GFlowNet policy to generate prompts proportional to reward.}
    \label{fig:overview}
    \vspace{-12pt}
\end{figure*}

In this section, we present \textbf{P}rompt \textbf{A}daptation with \textbf{G}FlowNets (\textbf{PAG)}, a novel framework that reformulates prompt adaptation as probabilistic inference through GFlowNets-based language model (LM) fine-tuning. 
We begin by outlining how we can fine-tune LMs with GFlowNets to satisfy our objective. Next, we investigate the critical challenge of mode collapse in this framework. To systematically address this challenge, we present our key technical advances that enhance GFlowNets for prompt adaptation. \Cref{fig:overview} illustrates the overview of our method.

\subsection{Problem Formulation}\label{sec:mode_collapse}
Unlike previous RL-based approaches like Prompist~\cite{hao2024optimizing}, which maximize the reward proxy and often converge to deterministic behaviors, we formulate prompt adaption as a probabilistic inference problem~\citep{zhao2024probabilistic} based on GFlowNets (as introduced in Section~\ref{sec:bg}). This formulation enables us to learn a policy that samples high-quality prompts while preserving the coverage of reference policy. Specifically, the optimal policy for Eq. \eqref{eq:objective} can be analytically derived as follows:
\begin{align}\label{eq:reward}
p_{\theta}^{*}(\mathbf{y}\vert\mathbf{x})\propto p_{\text{ref}}(\mathbf{y}\vert\mathbf{x})\cdot\exp\left(\frac{1}{\beta}r(\mathbf{x}, \mathbf{y})\right)=:R(\mathbf{x}, \mathbf{y}).
\end{align}
This formulation reduces prompt adaptation to learning how to sample from the unnormalized density function defined by rewards $R$, aiming to match the underlying reward distribution rather than solely maximizing the reward. A GFlowNet with such reward specification will learn a stochastic policy that generates prompts proportional to their rewards, thereby ensuring both effectiveness and diversity. Please refer to \cref{app:diff_btw_rl_and_gfn} for more discussion between RL and GFlowNets.

\begin{figure}[t]
    \centering
    \includegraphics[width=\linewidth]{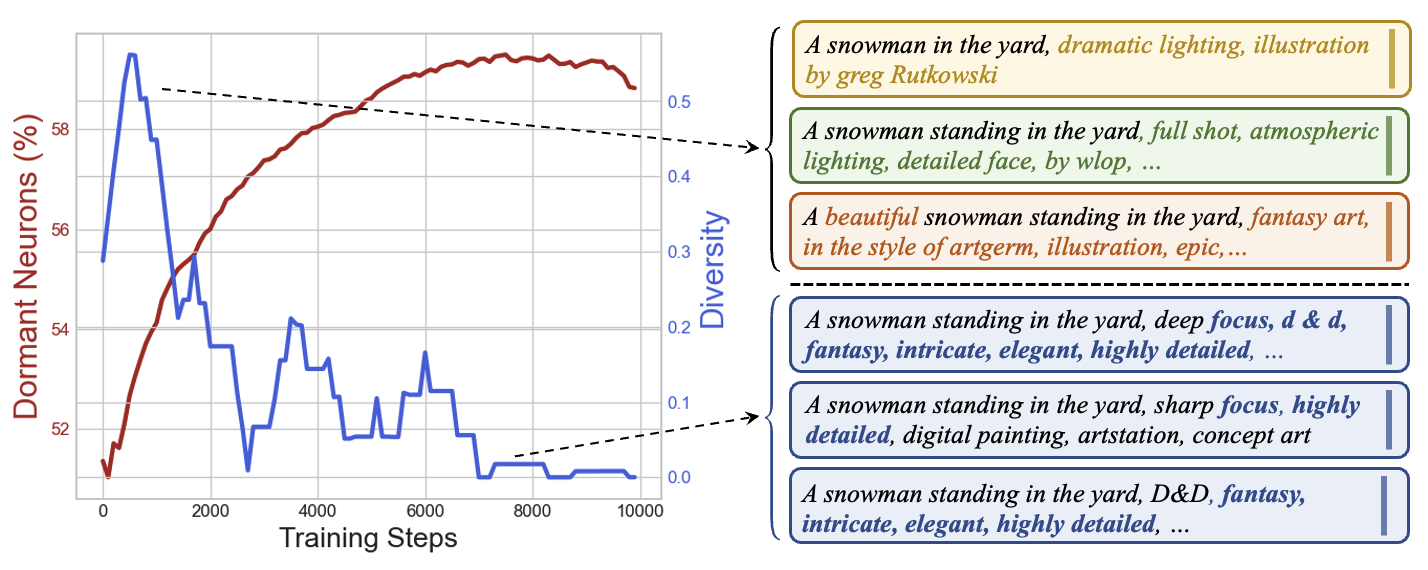}
    \caption{Mode collapse issue in prompt adaptation with a naive application of GFlowNets. The proportion of dormant neurons steadily increases (left), while the diversity of generated prompts significantly decreases over training iterations (right).}
    \label{fig:summary_mode_collapse}
    \vspace{-12pt}
\end{figure}

\vspace{5pt}
\noindent\textbf{Mode Collapse Issue in GFlowNets Fine-tuning}
While GFlowNets are designed to sample from the reward distribution for discovering high-quality and diverse candidates, our empirical analysis reveals a significant challenge in prompt adaptation, specifically when fine-tuning language models with GFlowNets after the supervised fine-tuning stage: the policy consistently generates similar prompts despite the vast token space as training progresses.
As shown in \Cref{fig:summary_mode_collapse}, directly applying GFlowNets for fine-tuning language models leads to a substantial reduction in the diversity of prompts. This outcome contradicts the original objective of GFlowNets, suggesting underlying limitations in the training dynamics.

To systematically diagnose the mode collapse issue, we investigate the learning behavior and neural activation patterns in the flow model during fine-tuning. By tracking the dormant neuron ratio as defined in \cref{eq:dormant}, we uncover a key finding largely overlooked in previous research -- the proportion of dormant neurons consistently increases throughout training, as illustrated in the left part in Figure~\ref{fig:summary_mode_collapse}. 
This progressive loss of neural plasticity largely constrains the capacity of the model to learn from and adapt to diverse regions in the prompt space. As more neurons become inactive, the model becomes inflexible in exploring the prompt space, leading to the generation of highly similar prompts.

The mode collapse is further exacerbated by inherent learning challenges in the sequential prompt generation process, where GFlowNets are primarily guided by terminal rewards that are only available after completing entire adapted prompts. It leads to inefficient credit assignment as the model struggles to attribute these terminal rewards into individual tokens in the generation process. Without clear feedback on which intermediate choices contribute to successful outcomes, the model tends to conservatively exploit discovered high-reward patterns rather than exploring diverse alternatives, as evidenced by the increasingly similar prompt patterns shown in Figure~\ref{fig:summary_mode_collapse}. 
These two challenges create a self-reinforcing cycle: the loss of plasticity limits the ability of GFlowNets to learn from diverse samples, while inefficient credit assignment hinders effective exploration of diverse alternatives, ultimately leading to the generation of increasingly similar prompt patterns and exacerbates the mode collapse.

These insights motivate our development of novel training mechanisms tailored for prompt adaptation with GFlowNets, which we detail in the following sections. 

\subsection{Proposed Method}
Based on our observations shown in \Cref{fig:summary_mode_collapse}, we recognize that the mode collapse problem emerges when GFlowNets struggles to maintain expressive capacity while learning from solely on terminal reward in the sequential process.

To systematically address these challenges, we propose a comprehensive mechanism that encourages GFlowNets to retain expressivity targeting three complementary aspects: i) flow reactivation strategy that maintains the plasticity of neural network for exploring diverse prompt patterns, ii) reward-prioritized sampling that allows the model to focus on high-reward experiences, and iii) decomposed reward structure that provides fine-grained learning signals throughout the generation process with advanced training guidance, 
which helps the model explore diverse prompts and resist the tendency toward mode collapse. 
The overall training framework is summarized in Algorithm~\ref{alg:main}.

\vspace{5pt}
\noindent\textbf{Flow Reactivation}
It has been proven that periodic network reset is effective in RL for maintaining neural plasticity~\citep{nikishin2022primacy, zhangconfronting, dsample, liu2024neuroplastic}. Drawing inspiration from this approach, we introduce a targeted flow reactivation mechanism that reinitializes only the last layer of the flow function periodically every $M$ steps.
Note that we do not reset the parameters of the forward policy, which directly interacts with the environment for sampling prompts, as it could cause drastic changes that destabilize training.

\vspace{5pt}
\noindent\textbf{Reward-Prioritized Sampling}
However, relying solely on reset strategies~\citep{nikishin2022primacy} encounter significant challenges in prompt adaptation due to the vast token space, which requires the model to undergo extensive re-exploration to re-discover previously identified high-reward regions, and can therefore impact training efficiency.

To maintain the expressivity of flow function while preventing a significant drop in performance that leads to instability, we leverage reward-prioritized sampling during off-policy training of GFlowNets \citep{shen2023towards, kim2024adaptive}.
A key insight of our approach lies in maintaining consistent access to high-quality prompts through prioritized sampling, which serves as a warm-up for accelerating high-quality knowledge recovery. 
Specifically, we sample a batch of prompts from the replay buffer with probabilities proportional to their rewards as follows:
\begin{align}\label{eq:prt}
    (\mathbf{x, y})\sim P_{\mathcal{B}}(\mathbf{x}, \mathbf{y})=\frac{\exp\left(R(\mathbf{x}, \mathbf{y})\right)}{\sum_{(\mathbf{x}, \mathbf{y})\in\mathcal{B}}\exp\left(R(\mathbf{x}, \mathbf{y})\right)}.
\end{align}
This reward-prioritized sampling naturally complements the flow reactivation process. While reset maintains the expressivity of the model, the continuous exposure to high-reward prompts ensures efficient knowledge retention. Through this principled approach, we can prevent the GFlowNet policy from forgetting the previously discovered high-scoring regions and can quickly regain its ability to navigate towards promising regions in the token space.

\vspace{5pt}
\noindent\textbf{Reward Decomposition}
Beyond maintaining model plasticity and high-quality data retention, a critical challenge in the mode collapse problem in prompt adaption lies in the inefficient credit assignment problem~\citep{pan2023better, jang2024learning} as discussed in Section~\ref{sec:mode_collapse}.
To address this issue, we propose a progressive reward decomposition scheme with advanced learning objectives. By carefully analyzing our reward function, we find that the likelihood term $p_{\text{ref}}(\mathbf{y}\vert\mathbf{x})$ in the reward function naturally admits a step-wise decomposition.
In other words, we can precisely extend the domain of our reward function from the set of terminal states to all possible states:
\begin{align}
    R(\mathbf{x}, y_{0:t})=
    \begin{dcases}
        p_{\text{ref}}(y_{0:t}\vert \mathbf{x}) & \text{if }t\neq T \\
        p_{\text{ref}}(\mathbf{y}\vert \mathbf{x})\exp\left(\frac{1}{\beta}r(\mathbf{x}, \mathbf{y})\right) & \text{otherwise}
    \end{dcases}
\end{align}
This enables us to extend the FL-DB objective in~\cref{eq:fl-db-objective}  by incorporating local credit signals at each step by minimizing the loss $\mathcal{L}(\mathbf{x},\mathbf{y};\theta)=\sum_{t=0}^{T-1}\mathcal{L}(\mathbf{x},\mathbf{y}_{0:t+1};\theta)$, where 
\begin{align}\label{eq:fl_db}
    &\mathcal{L}(\mathbf{x},\mathbf{y}_{0:t+1};\theta)\\
    &=\big(\log\tilde{F_{\theta}}(y_t\vert\mathbf{x}, y_{0:t-1})+\log P_{F}(y_{t+1}\vert\mathbf{x}, y_{0:t};\theta)\nonumber \\
    &+\log R(\mathbf{x}, y_{0:t}) - \log\tilde{F_{\theta}}(y_{t+1}\vert\mathbf{x}, y_{0:t})-\log R(\mathbf{x}, y_{0:t+1})\big)^2,\nonumber
\end{align}
with $F_{\theta}(y_{t+1}\vert\mathbf{x}, y_{0:t})=\tilde{F}_{\theta}(y_{t+1}\vert\mathbf{x}, y_{0:t})R(\mathbf{x}, y_{0:t})$. This enables the model to effectively assess token-level decisions, leading to an increase of diversity and mitigating mode collapse, as demonstrated in Section~\ref{sec:main_res}.

\begin{algorithm}[t]
    \caption{Prompt Adaptation with GFlowNets (PAG)}
    \algrenewcommand\algorithmicrequire{\textbf{Input:}}
    \algrenewcommand\algorithmicensure{\textbf{Output:}}
    \begin{algorithmic}[1]
    \Require{
        Initial prompt dataset $\mathcal{D}$, pretrained LM $p_{\text{ref}}$, number of training rounds $N$, reset period $M$, batch size $b$.
    }
    \Ensure{Fine-tuned policy $P_{F}(\cdot\vert\cdot;\theta)$}
    \State{Initialize $P_{F}(\cdot\vert\cdot;\theta)\leftarrow p_{\text{ref}}$, $\mathcal{B}\leftarrow\emptyset$}, and $F_{\theta}$
    \For {$n=1,\cdots,N$}
        \State{Initialize $\ell\leftarrow0$}
        \For {$i=1,\cdots,b$}
            \State{Uniformly sample $s$ from $\{0, 1\}$}
            \If{$s==0$} 
                \State{Sample $\mathbf{x}\sim\mathcal{D}$ and $\mathbf{y}^{i}\sim P_{F}(\cdot\vert\mathbf{x})$}
                \State{Compute $R(\mathbf{x},\mathbf{y}^{i})$ with \cref{eq:reward}}
                \State{$\mathcal{B}\leftarrow\mathcal{B}\cup\{(\mathbf{x}, \mathbf{y}^{i}, R(\mathbf{x}, \mathbf{y}^{i}))\}$}
            \Else
                \State{Sample $(\mathbf{x}, \mathbf{y}^{i}, R(\mathbf{x},\mathbf{y}^{i}))\sim\mathcal{B}$ with \cref{eq:prt}}
            \EndIf
            \State{Compute $\ell\leftarrow\ell+\mathcal{L}(\mathbf{x},\mathbf{y}^{i};\theta) / b$} with \cref{eq:fl_db}
        \EndFor
        \State{Update $P_{F}(\cdot\vert\cdot;\theta)$ and $F(\cdot;\theta)$} with computed loss $\ell$
        \If{$n\text{ mod }M == 0$}
            \State{Reset the last layer of $F(\cdot;\theta)$}
        \EndIf
    \EndFor
    \end{algorithmic}
    \label{alg:main}
\end{algorithm}

\section{Experiments}
In this section, we conduct extensive experiments to validate the effectiveness of our method on prompt adaptation. For the policy model, we use a pretrained GPT-2 \citep{radford2019language} following the setup in~\citep{hao2024optimizing}. As a default setting, we employ Stable-Diffusion v1.4 \citep{rombach2022high} as the target text-to-image model and use DPM solver \citep{lu2022dpm} with 20 inference steps to accelerate the sampling process. Detailed information regarding the experimental settings can be found in \Cref{app:exp_details}.

\begin{figure*}[t]
    \centering
    \includegraphics[width=\textwidth]{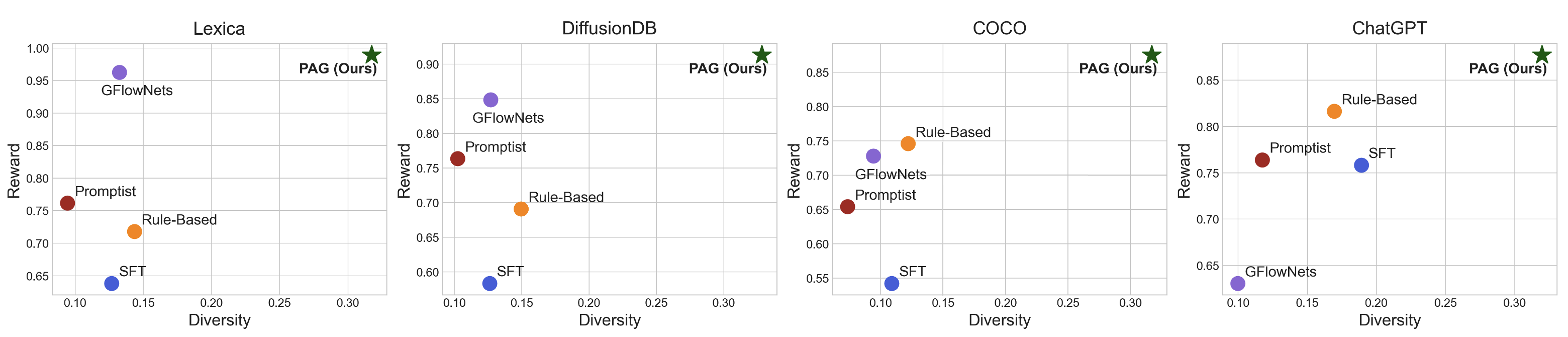}
    \caption{Reward and diversity of prompts generated by each method with different initial prompt datasets.}
    \label{fig:main_tradeoff}
    \vspace{-10pt}
\end{figure*}

\subsection{Experiment Setup}
\noindent\textbf{Dataset Preparation}
We strictly follow the setup of Promptist~\citep{hao2024optimizing} for dataset preparation. For training, we collect prompts from the Lexica website \citep{lexica}, DiffusionDB \citep{wang2023diffusiondb}, and COCO dataset \citep{chen2015microsoft}. For evaluation, we randomly sample $256$ prompts from the three types of training datasets. In addition, we introduce a challenging dataset using ChatGPT \citep{ouyang2022training} to
generate brief prompts that describe images around $5$ words. These brief prompts
naturally require diverse adaptations for better generation quality, representing a practical scenario for prompt adaptation.

\vspace{5pt}
\noindent\textbf{Baselines}
We consider several strong baselines to verify the efficacy of our method on the prompt adaptation task.
\begin{itemize}
    \item \textbf{Supervised Fine-tuning (SFT)}: A policy model fine-tuned by supervised learning on a set of prompt pairs of original user inputs and manually engineered prompts. 
    \item \textbf{Promptist}: A PPO-based approach \citep{schulman2017proximal} that directly trains the RL policy to maximize the reward function. 
    \item \textbf{Rule-Based}: Based on the observation that Promptist mostly generates similar postfixes with deterministic behaviors, we build a heuristic that appends the most frequently used postfixes in Promptist to the initial prompt.
    \item \textbf{GFlowNets}: We implement a vanilla GFlowNets method with the trajectory balance (TB) objective~\citep{malkin2022trajectory} to train the target policy. As our task is a conditional generation task, we use the VarGrad \citep{richter2020vargrad} version of TB loss, which is widely used for reducing variance \citep{zhangrobust, kim2024ant}.
    \item \textbf{DPO-Diff}: A gradient-based optimization method designed to discover effective negative prompts \citep{wang2024discrete}. As it is not directly comparable, we provide more discussion on DPO-Diff in Appendix~\ref{app:extend_main_results}. 
\end{itemize}

\vspace{5pt}
\noindent\textbf{Training and Evaluation}
Following Promptist~\citep{hao2024optimizing}, we initialize the policy with SFT policy before GFlowNet fine-tuning. To parametrize the flow function, we use a separate neural network that takes the last hidden embedding of the prompts as input and outputs a scalar value. We train both the policy and flow function for 10K steps with a batch size of 256. For learning rate, we use $1\times10^{-5}$ for the policy and $1\times10^{-4}$ for the flow function.

For evaluation, we generate $16$ prompts for each prompt via beam search with a length penalty of $1.0$. Then we generate $3$ images per prompt to compute the reward. To measure diversity, we compute the average pairwise cosine distance of 16 prompts for each initial prompt.

\subsection{Performance Comparison} \label{sec:main_res}
\Cref{fig:main_tradeoff} presents a comprehensive comparison of PAG against the baselines across different datasets. As depicted in the figure, while Promptist improves upon SFT through RL-based fine-tuning, it exhibits limited diversity due to the reward-maximization nature.
Vanilla GFlowNets mostly achieves significantly higher rewards than both SFT and Promptist, but still suffers from mode collapse. In contrast, PAG consistently surpasses all baselines in terms of both reward and diversity metrics, demonstrating its capability to generate both high-quality and diverse prompts across various input types.

\begin{figure*}[t]
    \centering
    \includegraphics[width=0.95\textwidth]{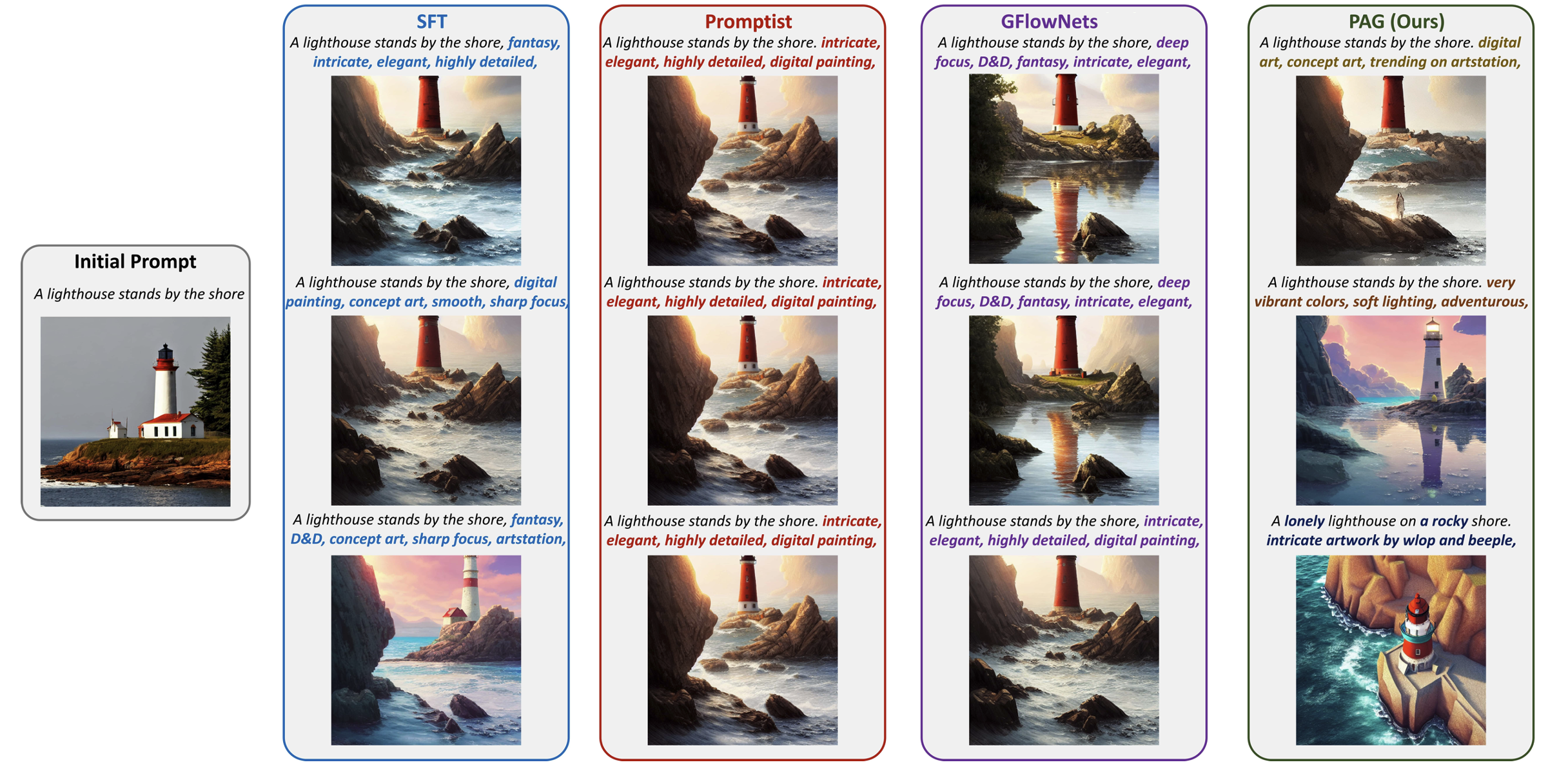}
    \caption{Images generated by optimized prompts using Stable Diffusion v1.4 (with the same seed to visualize the effect solely on prompts).
    Our method generates diverse and highly aesthetic images based on adapted prompts of high quality and diversity.}
    \label{fig:main_figure}
    \vspace{-12pt}
\end{figure*}

The challenging ChatGPT dataset, where initial prompts usually give us little information, further highlights the strengths of our approach. 
We observe that Promptist or vanilla GFlowNets lead to severe mode collapse and are outperformed by simple deterministic rule-based heuristics, while our method maintains robust performance.
We further visualize the generated prompts and corresponding images from each method in \Cref{fig:main_figure} (more figures can be found in Appendix~\ref{app:more_visualization}). As illustrated in the figures, PAG generates diverse and meaningful enhancements by not only appending postfixes but also introducing relevant adjectives and detailed descriptions, which is particularly valuable when the original prompt lacks sufficient information.

\begin{table}[t]
\centering
\caption{Reward and diversity of prompts generated by each method with different reward functions.}
\vspace{-5pt}
\resizebox{\linewidth}{!}{
\begin{tabular}{lcc|cc}
\toprule
\multirow{2}{*}{\textbf{Method}} & \multicolumn{2}{c}{ImageReward} & \multicolumn{2}{c}{HPSv2}  \\
\cmidrule{2-5}
& Reward & Diversity & Reward & Diversity \\
\midrule
SFT & 0.66 ± 0.01 & 0.14 ± 0.00 & 0.01 ± 0.00 & 0.14 ± 0.00 \\
Promptist & 0.70 ± 0.05 & 0.02 ± 0.00 & -0.05 ± 0.01 & 0.03 ± 0.00 \\
Rule-Based & 0.59 ± 0.02 & 0.12 ± 0.00 & 0.01 ± 0.00 & 0.12 ± 0.00 \\
GFlowNets & 0.81 ± 0.15 & 0.20 ± 0.00 & \textbf{0.02 ± 0.00} & 0.14 ± 0.00 \\
\midrule
\textbf{PAG (Ours)} & \textbf{0.83 ± 0.01} & \textbf{0.29 ± 0.00} & \textbf{0.02 ± 0.00} & \textbf{0.37 ± 0.00} \\
\bottomrule
\end{tabular}}
\label{tab:extend_reward_fn}
\vspace{-10pt}
\end{table}
\vspace{-.03in}
\subsection{Robustness across Different Reward Functions}
To evaluate the robustness of our framework in text-to-image modeling, we extend our experiments to include 
two additional widely-used reward functions in diffusion alignment: ImageReward \citep{xu2024imagereward} and HPSv2 \citep{wu2023human}. We maintain the same training procedure as our primary experiments while incorporating these different reward functions, and utilize COCO dataset for evaluation. Please refer to \Cref{app:exp_details} for more detailed experimental procedures. 

\Cref{tab:extend_reward_fn} summarizes the results of different methods, which demonstrates that PAG achieves state-of-the-art performance in terms of both reward and diversity metrics. This consistent performance across multiple reward functions validates the robustness of our approach. 

\subsection{Transferability to Different Text-to-Image Diffusion Models}
Since our method operates on prompt adaptation without modifying the underlying diffusion model parameters,
it has the potential to generalize across different text-to-image diffusion models in a zero-shot manner. 
To validate this, we evaluate the transferability of our method on various representative text-to-image diffusion models distinct from the target model used during training, including SD v1.5 \citep{rombach2022high}, SSD-1B \citep{gupta2024progressive}, SDXL-Turbo \citep{sauer2025adversarial}, and SD3 \citep{esser2024scaling} for evaluation. We use prompts generated by each method using SD v1.4 with initial prompts from the ChatGPT dataset.

\Cref{tab:transfer_t2i_models} summarizes the performance of various methods on different text-to-image diffusion models. As demonstrated in the table, PAG consistently generates high-rewarding images compared to other methods due to its ability to produce diverse prompts, which provides robustness on different text-to-image diffusion models. Surprisingly, we observe that there is a significant gap between our method and baselines in the SD3 model, showcasing that our method can be applied in practical settings. We also visualize the generated images across different text-to-image diffusion models in \cref{app:more_visualization}.

\begin{figure*}[t]
    \centering
    \includegraphics[width=0.95\textwidth]{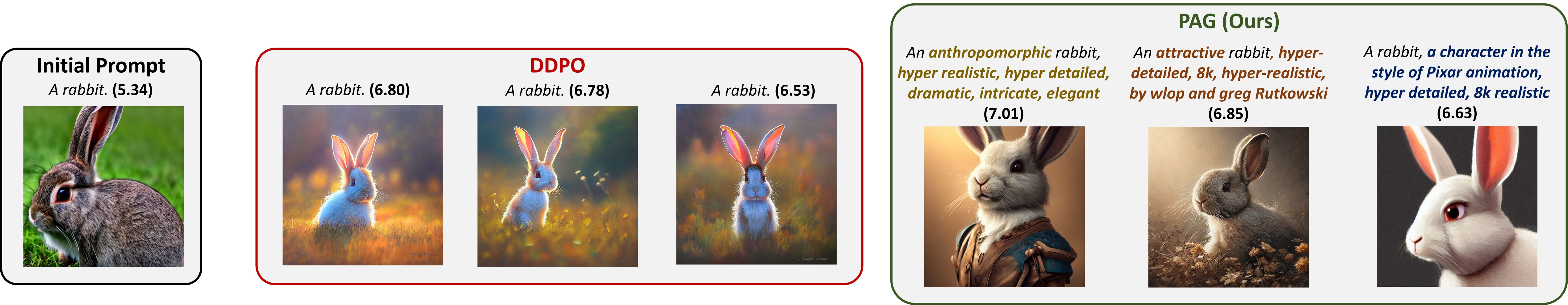}
    \caption{Comparison with DDPO and PAG. We report the aesthetic score of images in bold.}
    \label{fig:ddpo_comparison}
    \vspace{-12pt}
\end{figure*}
\begin{table}[t]
\centering
\caption{We train the policy with SD v1.4 as a target model and evaluate the generated propmts with different text-to-image models in a zero-shot manner.}
\vspace{-5pt}
\resizebox{\linewidth}{!}{
\begin{tabular}{lcccc}
\toprule
\multirow{2}{*}{\textbf{Method}} & \multicolumn{4}{c}{Text-to-Image Diffusion Models} \\
\cmidrule{2-5}
& SD v1.5 & SSD-1B & SDXL-Turbo & SD3 \\
\midrule
SFT & 0.78 ± 0.05 & 0.53 ± 0.05 & 0.54 ± 0.05 & 0.77 ± 0.04 \\
Promptist & 0.80 ± 0.03 & 0.46 ± 0.05 & 0.47 ± 0.06 & 0.73 ± 0.03 \\
Rule-Based & 0.85 ± 0.02 & 0.54 ± 0.05 & 0.54 ± 0.05 & 0.76 ± 0.03 \\
GFlowNets & 0.62 ± 0.04 & 0.51 ± 0.02 & 0.49 ± 0.01 & 0.75 ± 0.06 \\
\midrule
\textbf{PAG (Ours)}  & \textbf{0.87 ± 0.02} & \textbf{0.61 ± 0.04} & \textbf{0.67 ± 0.02} & \textbf{0.95 ± 0.05} \\
\bottomrule
\end{tabular}}
\label{tab:transfer_t2i_models}
\vspace{-15pt}
\end{table}

\subsection{Comparison with Fine-tuning Text-to-Image Diffusion Models}
To verify the effectiveness of our framework, we also compare our method with approaches that directly fine-tune text-to-image diffusion models, and compare with DDPO \citep{blacktraining}, which is trained with the aesthetic score reward function on animal prompts. As shown in \Cref{fig:ddpo_comparison}, we find that PAG achieves competitive performance with DDPO in terms of aesthetic quality. Furthermore, we observe that generated samples from DDPO converge to similar styles, whereas PAG generates diverse and high-quality images, indicating that prompt adaptation can be a promising alternative and a complementary approach to directly fine-tuning diffusion models for generating images with desired properties.  
For more details on comparison between directly fine-tuning diffusion models and our method, please refer to \Cref{app:comp_diffusion}.

\subsection{Ablation Studies}
In this section, we conduct comprehensive ablation studies to investigate the effectiveness of the important components and analyze how our method systematically tackles the severe mode collapse problem. 

\Cref{tab:ablation} shows the effectiveness of flow reactivation (Reset), reward-prioritized sampling (PRT), and reward decomposition (FL) in COCO and ChatGPT datasets. As shown, each component contributes substantially to the overall performance. Notably, we observe that removing PRT leads to significant performance degradation, which indicates that we should carefully tackle the mode collapse issue from the angle of both network parameters and training samples. Moreover, reward decomposition improves both reward and diversity metrics, validating its effectiveness for better credit assignment and mitigating mode collapse.
Additional component analysis can be found in Appendix~\ref{app:extend_ablation}.

\begin{table}[t]
\centering
\caption{Ablation studies on each component of our method.}
\vspace{-5pt}
\resizebox{\linewidth}{!}{
\begin{tabular}{lcc|cc}
\toprule
\multirow{3}{*}{\textbf{Method}} & \multicolumn{2}{c}{COCO} & \multicolumn{2}{c}{ChatGPT}  \\
\cmidrule{2-5}
& Reward & Diversity & Reward & Diversity \\
\midrule
Ours & 0.88 ± 0.02 & 0.32 ± 0.00 & 0.88 ± 0.04 & 0.32 ± 0.00 \\
\midrule
w/o Reset & 0.70 ± 0.00 & 0.27 ± 0.00 & 0.81 ± 0.05 & 0.31 ± 0.00 \\
w/o PRT   & 0.30 ± 0.01 & 0.22 ± 0.00 & 0.72 ± 0.08 & 0.23 ± 0.00 \\
w/o FL    & 0.56 ± 0.01 & 0.16 ± 0.00 & 0.81 ± 0.02 & 0.18 ± 0.00\\
\bottomrule
\end{tabular}}
\label{tab:ablation}
\vspace{-12pt}
\end{table}
\begin{figure}[t]
\begin{minipage}[t]{\linewidth}
    \begin{subfigure}[t]{0.48\linewidth}
        \centering
        \includegraphics[width=\textwidth]{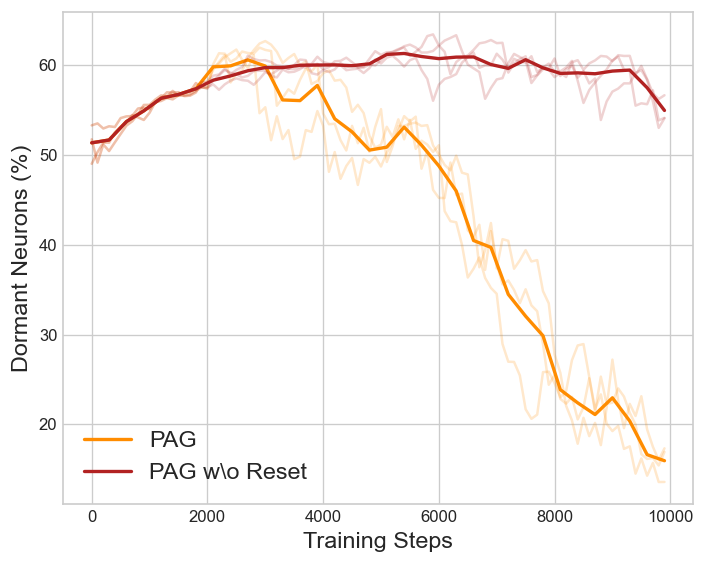}
    \end{subfigure}
    \begin{subfigure}[t]{0.48\linewidth}
        \centering
        \includegraphics[width=\textwidth]{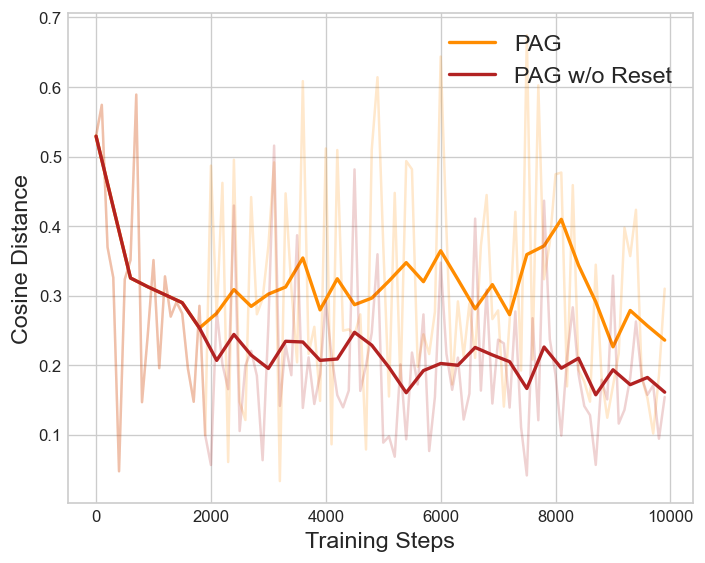}
    \end{subfigure}
    \caption{The effect of flow reactivation.}
    \label{fig:ablation}
\end{minipage}
\vspace{-15pt}
\end{figure}

Furthermore, we carefully analyze how flow reactivation mitigates the mode collapse problem by tracking the percentage of dormant neurons (based on Eq.~\ref{eq:dormant}) over the course of training. As depicted in \Cref{fig:ablation}, our flow reactivation mechanism exhibits a significantly lower dormant neuron rate, while the variant without this mechanism results in a high portion of dormant neurons during training. We also plot the cosine distance among generated prompts over training and observe that, without reactivation, the GFlowNet policy suffers from the mode collapse issue.
These findings validate its effectiveness in preventing our GFlowNets-based agent from converging to limited patterns by maintaining the expressivity of the flows, contributing to both performance and diversity improvements.

\section{Related Works}
\noindent\textbf{Aligning Diffusion Models}
There is a surge of interest in generating images with desired properties, which can be modeled as reward functions from human feedback \citep{ouyang2022training}. 
A widely recognized approach involves directly fine-tuning diffusion models with the reward function. \citet{blacktraining} and \citet{fan2024reinforcement} formulate the diffusion reverse process as a MDP and employ RL for fine-tuning diffusion models while \citet{clarkdirectly} and \citet{prabhudesai2023aligning} update model parameters through end-to-end backpropagation of the gradient of reward across denoising steps. 
While those methods have shown promising results, 
they require initiating fine-tuning from scratch for each different text-to-image diffusion model and necessitate access to the model parameters, which may be restricted due to confidential issues \citep{ramesh2022hierarchical, saharia2022photorealistic}. Unlike these methods, PAG enhances initial prompts into high-quality and diverse prompts by fine-tuning language models, allowing transferability across various text-to-image models. 

\vspace{5pt}
\noindent\textbf{Prompt Adaptation for Text-to-Image Models}
There have been some trials to generate high-quality images by adapting prompts instead of fine-tuning text-to-image models. A pioneering work of this approach is Promptist \citep{hao2024optimizing}, which formulates prompt adaptation as an RL problem.
A relevant recent work, DPO-Diff, \citep{wang2024discrete}, also tries to generate user-aligned images while optimizing negative prompts using a shortcut text gradient. We find that Promptist often results in deterministic policy, which can be easily replaced by heuristics. PAG utilizes GFlowNets to fine-tune language models for generating effective and diverse prompts. 

\vspace{5pt}
\noindent\textbf{GFlowNet Fine-Tuning}
GFlowNets are probabilistic methods that sample compositional objects proportional to unnormalized density through sequential decision-making~\citep{bengio2021flow, bengio2023gflownet} and energy-based modeling~\citep{zhang2022generative}, with applications in structure learning~\citep{deleu2022bayesian} and combinatorial optimization~\citep{zhangrobust,Zhang2023LetTF}, which can also be extended to continuous~\citep{lahlou2023cgfn} or stochastic scenarios~\citep{pan2023stochastic,zhang2023distributional}.
It has the potential for fine-tuning language models (LMs) for intractable posterior inference problems~\citep{huamortizing} and robust red-teaming~\citep{lee2024learning}.
Although there have been several attempts in improving exploration and training efficiency~\citep{pan2023generative, kimlocal, lau2024qgfn} and extending it to more general domains and learning paradigms, previous works typically train a GFlowNets policy from scratch~\citep{pan2024pre,he2024looking}, and largely overlooked the critical problem of plasticity loss during fine-tuning~\citep{zhang2024improving,liu2024efficientdiversitypreservingdiffusionalignment}.
Furthermore, most prior methods focus on unconditional generation and often suffer from mode collapse, requiring an additional post-supervised fine-tuning stage. 
In this work, we investigate the mode collapse problem in prompt adaptation and propose PAG, a novel approach to address this key challenge for diverse conditional prompt generation for text-to-image diffusion models.

\section{Conclusion}
In this paper, we propose a novel approach that systematically addresses mode collapse in prompt adaptation through GFlowNets-based probabilistic inference. We identify a critical plasticity loss problem in GFlowNets-based prompt adaptation when learning from rewards sequentially, and present PAG to maintain generation flexibility. Our extensive results show that PAG successfully learns to sample effective and diverse prompts for text-to-image generation. 

{
    \bibliographystyle{ieeenat_fullname}
    \bibliography{main}
}

\clearpage
\appendix

\noindent Our code is publicly available \href{https://github.com/dbsxodud-11/PAG.git}{here}.

\section{Illustrations of RL and GFlowNets}\label{app:diff_btw_rl_and_gfn}
\begin{figure}[h]
    \centering
    \includegraphics[width=0.9\linewidth]{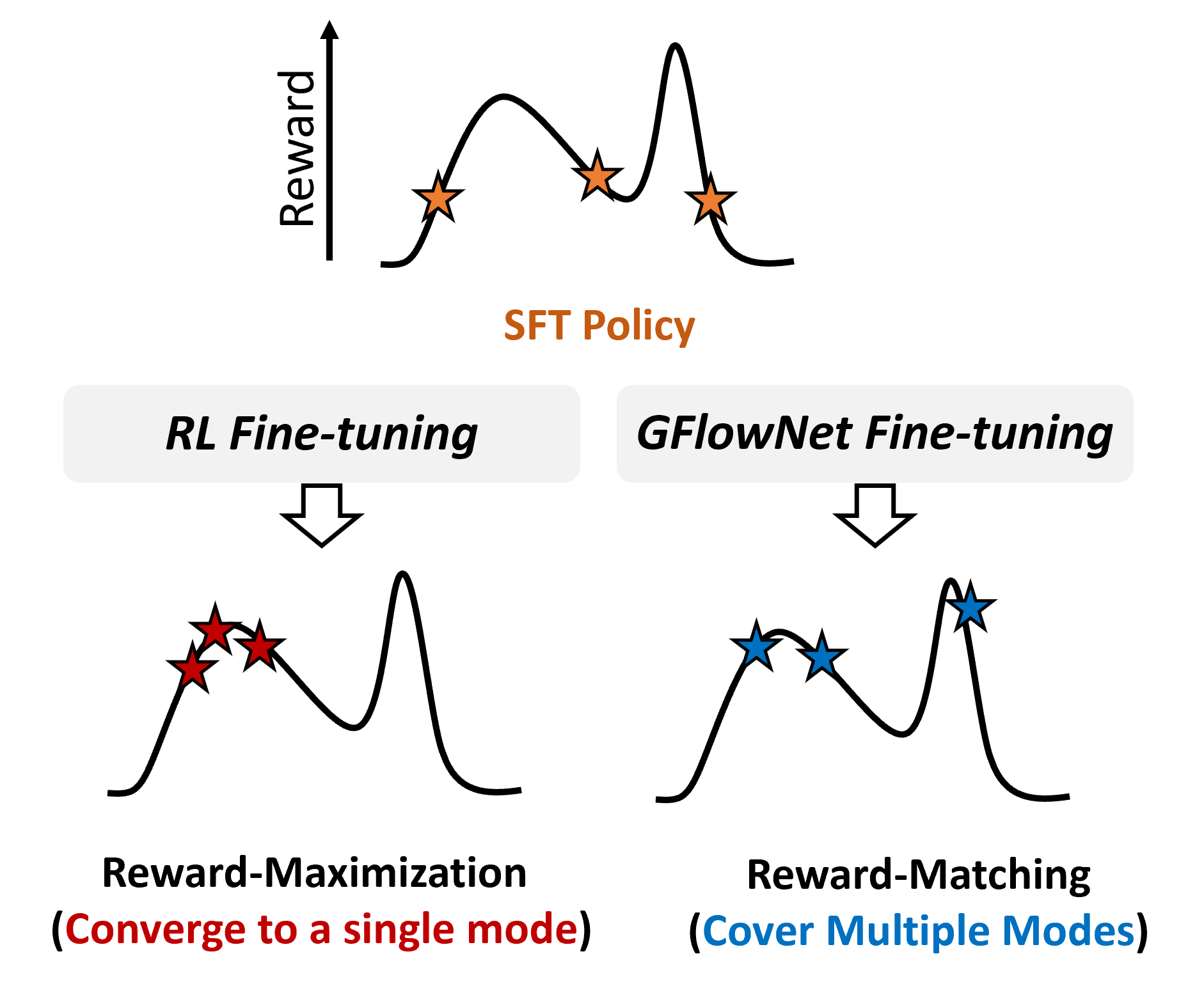}
    \vspace{-5pt}
    \caption{Comparison of the learning objective of reward-maximizing RL and reward-matching GFlowNets.}
    \label{fig:rl_vs_gfn}
    \vspace{-12pt}
\end{figure}

In this section, we summarize the key difference between Reinforcement Learning (RL) and GFlowNets in more detail. GFlowNets~\citep{bengio2021flow} is designed to learn a stochastic policy that generates samples proportional to their rewards (i.e., $p(x)\propto R(x)$), while RL aims to learn a policy that maximizes the reward function as illustrated in \Cref{fig:rl_vs_gfn}. Learning a policy that samples proportional to the reward function leads to capturing multi-modal distribution and discovering high-quality and diverse candidates, which is particularly beneficial when the reward proxy is inaccurate \citep{bengio2021flow}, and their connections have been studied in \citep{tiapkin2024generative,deleu2024discrete,he2024rectifying}, which shares equivalences and similarities with entropy-regularized RL~\citep{nachum2017bridging} in tree-structured sequence generation and directed acyclic graph problems~\citep{huamortizing,lee2024learning}.
In prompt adaptation tasks, conventional RL-based approaches~\citep{hao2024optimizing} based on PPO~\citep{schulman2017proximal} that naively maximize a reward function can lead to reward overoptimization and hinder generalizability to different text-to-image diffusion models, while it is more desirable to generate not only effective and but also diverse prompts.     

\section{Experiment Setting Details}\label{app:exp_details}
In this section, we present details of experiment settings.
\subsection{Data Preparation}
We strictly follow the procedure of Promptist \citep{hao2024optimizing} to prepare training and evaluation datasets. Additionally, we introduce a challenging dataset using ChatGPT \citep{ouyang2022training} interface to generate brief prompts that describe images. Specifically, we use the following prompt to query ChatGPT for short image descriptions:
\begin{itemize}
    \item \texttt{Generate N sentences describing photos /pictures/images with length around 5.}
\end{itemize}

\noindent Below are a few example prompts generated by ChatGPT:
\begin{itemize}
    \item \texttt{A bird sitting on a branch.}
    \item \texttt{A tree under a sky.}
    \item \texttt{A car drives on a road.}
    \item \texttt{A train moves through the city.}
    \item \texttt{A boat sails on a lake.}
\end{itemize}

\subsection{Baselines}
In this section, we provide more details on the baselines used for our experiments.

\paragraph{Supervised Fine-Tuning.} We fine-tune the pretrained GPT-2 policy model with supervised learning on a set of prompt pairs of original user inputs and manually engineered prompts provided by Promptist \citep{hao2024optimizing}. As a default, we use the pretrained weights of SFT publicly available\footnote{https://github.com/microsoft/LMOps/tree/main/promptist}.

\paragraph{Promptist.} We train the policy with a PPO-based approach where the policy is initialized with the supervised fine-tuned model. As a default, we use the pretrained weights of Promptist publicly available to ensure a fair comparison. To evaluate the generalizability of different reward functions, we train the policy with the same hyperparameter configurations.

\paragraph{Rule-Based.} Based on the observation that Promptist mostly generates similar postfixes for optimization, we build a rule-based method that appends the most frequently used postfixes in Promptist to the initial prompt. Below are a few example postfixes we used for evaluation:
\begin{itemize}
    \item \texttt{intricate, elegant, highly detailed, digital painting, artstation, concept art, sharp focus, illustration, by justin gerard and artgerm, 8 k.}
    \item \texttt{by greg rutkowski, digital art, realistic painting, fantasy, very detailed, trending on artstation.}
    \item \texttt{highly detailed, digital painting, artstation, concept art, sharp focus, illustration, art by greg rutkowski and alphonse mucha.}
\end{itemize}

\paragraph{GFlowNets.} We train a vanilla GFlowNets policy with TB \citep{malkin2022trajectory} objective. As our task is a conditional generation task, a naive implementation of TB should directly estimate a conditional partition function, $Z_{\theta}(\mathbf{x})$, which makes learning highly unstable \citep{kim2024ant}. For pratical implementation, we use VarGrad \citep{richter2020vargrad} objective to fine-tune the policy, which is widely used for reducing variance in GFlowNets literature \citep{zhangrobust, venkatraman2024amortizing}.

For each initial prompt $\mathbf{x}$ sampled in the minibatch, we generate $k=16$ on-policy samples $\mathbf{y}^{(1)},\cdots,\mathbf{y}^{(k)}$ with the forward policy. Each sample can be used to implicitly estimate $\log Z(\mathbf{x})$:
\begin{align*}
    \log \hat{Z}(\mathbf{x})^{(i)}=R(\mathbf{x},\mathbf{y}^{(i)})-\sum_{t=1}^{T}\log P_{F}(y_{t}^{(i)}\vert y_{0:t-1}^{(i)},\mathbf{x};\theta)
\end{align*}
Then we minimize the sample variance across the minibatch as follows:
\begin{align*}
    \mathcal{L}(\mathbf{x};\theta)=\frac{1}{k}\sum_{i=1}^{k}\left(
    \log\hat{Z}(\mathbf{x})^{(i)}-\frac{1}{k}\sum_{j=1}^{k}\log\hat{Z}(\mathbf{x})^{(j)}\right)^2
\end{align*}

\paragraph{DPO-Diff.} \citet{wang2024discrete} proposed a discrete prompt optimization for diffusion models (DPO-Diff), which is a gradient-based optimization method for discovering effective negative prompts to generate user-aligned images. It first generates a compact subspace comprised of only the most relevant words to user input with ChatGPT API, then uses shortcut text gradients to efficiently compute the text gradient for optimization. As the original reward function of DPO-Diff is a spherical clip loss, we replace the reward function the same as \cref{eq:task_reward}. As we consider a setting where the reward function is a black-box function, we use the evolutionary search module suggested by the paper for a fair comparison. Please refer to the paper for more details.

\subsection{Training and Evaluation}
For training, we initialize the policy with the SFT policy before the GFlowNet fine-tuning. To parametrize the flow function, we use a separate neural network that takes the last hidden embedding of the prompts as input and outputs a scalar value. We conducted all experiments with 4 NVIDIA A100 GPUs, and the training took approximately 24 hours.

For evaluation, we study two metrics: reward and diversity. To compute the reward, we generate $N=16$ prompts for each initial prompt via beam search with a length penalty of $1.0$. Then, we generate three images per prompt to compute the reward. We aggregate the max score among $N$ prompts and compute the average across initial prompts. 
\begin{align*}
    \text{Reward}(\mathcal{D}_{\text{eval}})&:=\frac{1}{\vert\mathcal{D}_{\text{eval}}\vert}\sum_{\mathbf{x}\in\mathcal{D}_{\text{eval}}}\max_{\mathbf{y}\sim p_{\theta}(\cdot\vert\mathbf{x})}\left(r(\mathbf{x},\mathbf{y})\right)
\end{align*}
To compute diversity, we embed the generated prompts using MiniLMv2 \citep{wang2020minilmv2} encoder, compute the average pairwise cosine distance between embeddings of the prompts, and compute the average across initial prompts.
For all evaluations, we conduct experiments with four random seeds and report the mean and standard deviation.

The input format for both training and evaluation is \texttt{[Initial Prompt] Rephrase:}, following Promptist \citep{hao2024optimizing}. The hyperparameters we used for modeling and training are listed in \Cref{tab:hyperparam}. We conduct several ablations studies on the impact of various hyperparameters in \Cref{app:extend_ablation}.
\begin{table}[h]
\centering
\caption{Hyperparameters for Training PAG.}
\resizebox{0.9\linewidth}{!}{
\begin{tabular}{ll}
\toprule
Parameters & Values \\
\midrule
Batch size & 64 \\
Buffer size & 5000 \\
Optimizer & Adam \\
Training Steps & $1 \times 10^4$ \\
Learning Rate ($\gamma$) & $1 \times 10^{-5}$ (Policy), $1 \times 10^{-4}$ (Flow)\\
Temperature ($\beta$) & 0.05 \\
Reset Period ($M$) & 2000 \\
\bottomrule
\end{tabular}
}
\label{tab:hyperparam}
\end{table}

\subsection{Robustness across Different Reward Functions}
To evaluate the robustness of our framework in terms of different reward functions, we use two widely-used reward functions in diffusion alignment: ImageReward \citep{xu2024imagereward} and HPSv2 \citep{wu2023human}. We provide a detailed description of each function below.
\paragraph{ImageReward} ImageReward is a general-purpose text-to-image preference reward model trained with pairs of prompts and images. To compute the score, ImageReward extracts image and text embeddings using BLIP \cite{li2022blip} encoder and combines them with cross attention, and uses MLP to generate a scale value for preference comparison.
\paragraph{HPSv2} HPSv2 is a scoring model that accurately predicts human preferences on generated images. To accurately predict the score, it fine-tunes the CLIP \citep{radford2021learning} with the HPDv2 dataset, a large-scale dataset that captures human preferences on images from various sources.

\subsection{Transferability to Different Text-to-Image Diffusion Models}
To validate the transferability of our framework to different text-to-image diffusion models, we prepare several widely-used text-to-image diffusion models: SD v1.5 \citep{rombach2022high}, SSD-1B \citep{gupta2024progressive}, SDXL-Turbo \citep{sauer2025adversarial}, and SD3 \citep{esser2024scaling}. As SDXL-Turbo and SD3 models do not align with DPM solver \citep{lu2022dpm}, we use a standard generation pipeline, which uses a PNDM scheduler \citep{liu2022pseudo} with 20 inference steps. Furthermore, as SDXL-Turbo does not use the guidance scale, we set the guidance scale to 0. for SDXL-Turbo, and 7.5 (default) for others. 

\begin{table*}[t]
\centering
\caption{Performance of prompt generated by each method. Aes score indicates aesthetic quality improvement compared to the image generated with the original input. Experiments are conducted with four random seeds, and mean and standard deviation are reported. \textbf{Bold} represent the best entry.}
\vspace{-5pt}
\resizebox{0.95\textwidth}{!}{
\begin{tabular}{lcc|cc|cc|cc}
\toprule
\multirow{3}{*}{\textbf{Method}} & 
\multicolumn{2}{c}{Lexica} & \multicolumn{2}{c}{DiffusionDB} &
\multicolumn{2}{c}{COCO} & \multicolumn{2}{c}{ChatGPT}  \\
\cmidrule{2-9}
& Reward & Diversity & Reward & Diversity 
& Reward & Diversity & Reward & Diversity \\
\midrule
Initial Prompt & -0.16 ± 0.03 & - 
               & -0.22 ± 0.02 & -
               & -0.35 ± 0.01 & -
               & -0.42 ± 0.03 & - \\
\midrule
SFT & 0.64 ± 0.02 & 0.13 ± 0.00
    & 0.58 ± 0.01 & 0.13 ± 0.00 
    & 0.54 ± 0.07 & 0.11 ± 0.02
    & 0.76 ± 0.03 & 0.19 ± 0.00 \\
Promptist & 0.76 ± 0.02 & 0.09 ± 0.00
    & 0.76 ± 0.03 & 0.10 ± 0.00
    & 0.65 ± 0.02 & 0.07 ± 0.00 
    & 0.76 ± 0.03 & 0.12 ± 0.00 \\
Rule-Based & 0.72 ± 0.02 & 0.26 ± 0.00
    & 0.69 ± 0.02 & 0.15 ± 0.00 
    & 0.75 ± 0.01 & 0.12 ± 0.00
    & 0.82 ± 0.03 & 0.17 ± 0.00 \\
GFlowNets & 0.96 ± 0.01 & 0.13 ± 0.00
    & 0.85 ± 0.03 & 0.13 ± 0.00
    & 0.73 ± 0.02 & 0.09 ± 0.00
    & 0.63 ± 0.03 & 0.10 ± 0.00 \\
DPO-Diff & 0.13 ± 0.02 & -
    & 0.28 ± 0.06 & -
    & -0.03 ± 0.06 & -
    & -0.17 ± 0.03 & - \\
\midrule
PAG (Ours) & \textbf{0.99 ± 0.05} &  \textbf{0.32 ± 0.00} 
    & \textbf{0.91 ± 0.04} & \textbf{0.33 ± 0.00}
    & \textbf{0.88 ± 0.02} &  \textbf{0.32 ± 0.00} 
    & \textbf{0.88 ± 0.04} & \textbf{0.32 ± 0.00} \\
\bottomrule
\end{tabular}}
\label{tab:main1}
\end{table*}
\begin{figure*}[t]
\begin{minipage}[t]{\textwidth}
    \begin{subfigure}[t]{0.24\textwidth}
        \centering
        \includegraphics[width=\textwidth]{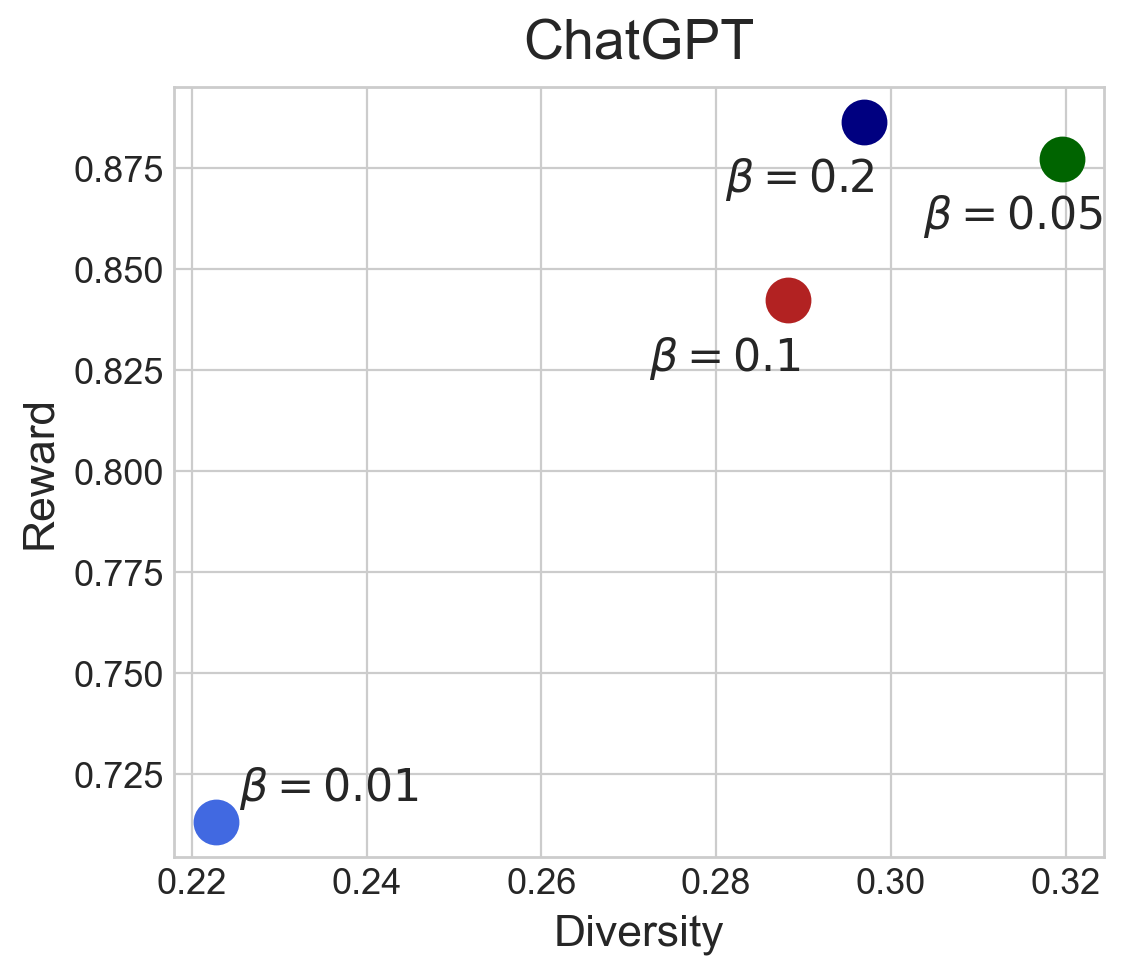}
        \subcaption{Analysis on $\beta$}
        \label{fig:ablation_beta}
    \end{subfigure}
    \begin{subfigure}[t]{0.24\textwidth}
        \centering
        \includegraphics[width=\textwidth]{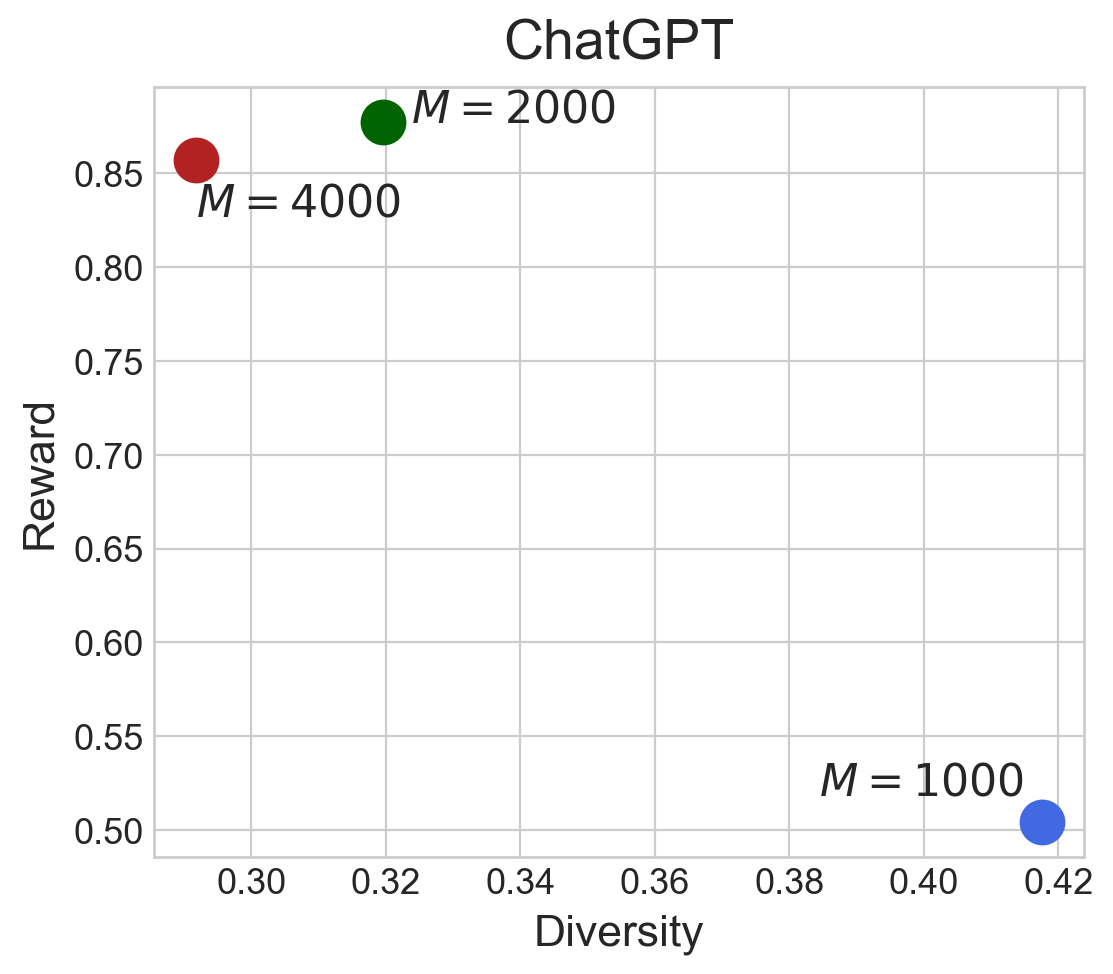}
        \subcaption{Analysis on $M$}
        \label{fig:ablation_M}
    \end{subfigure}
    \begin{subfigure}[t]{0.24\textwidth}
        \centering
        \includegraphics[width=\textwidth]{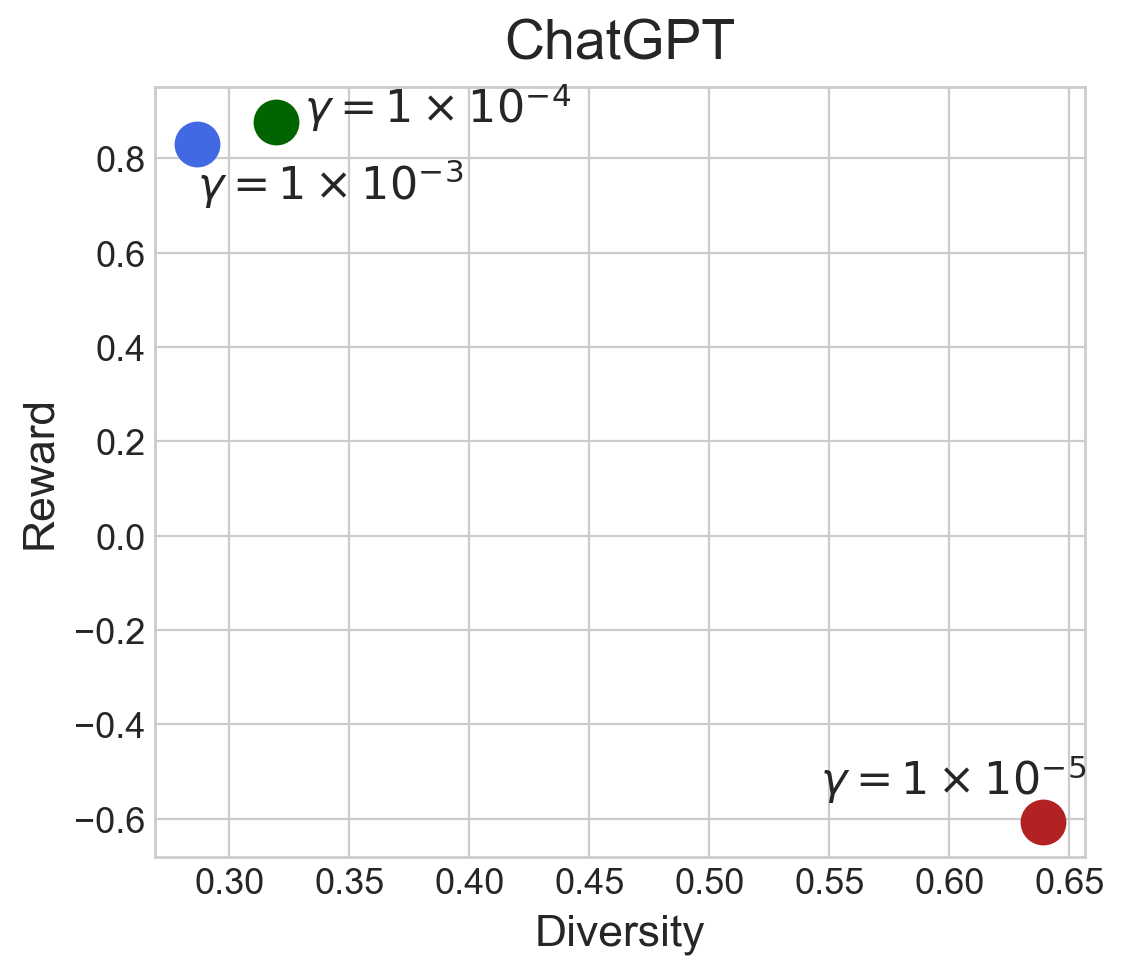}
        \subcaption{Analysis on $\gamma$ for flow function}
        \label{fig:ablation_gamma}
    \end{subfigure}
    \begin{subfigure}[t]{0.24\textwidth}
        \centering
        \includegraphics[width=\textwidth]{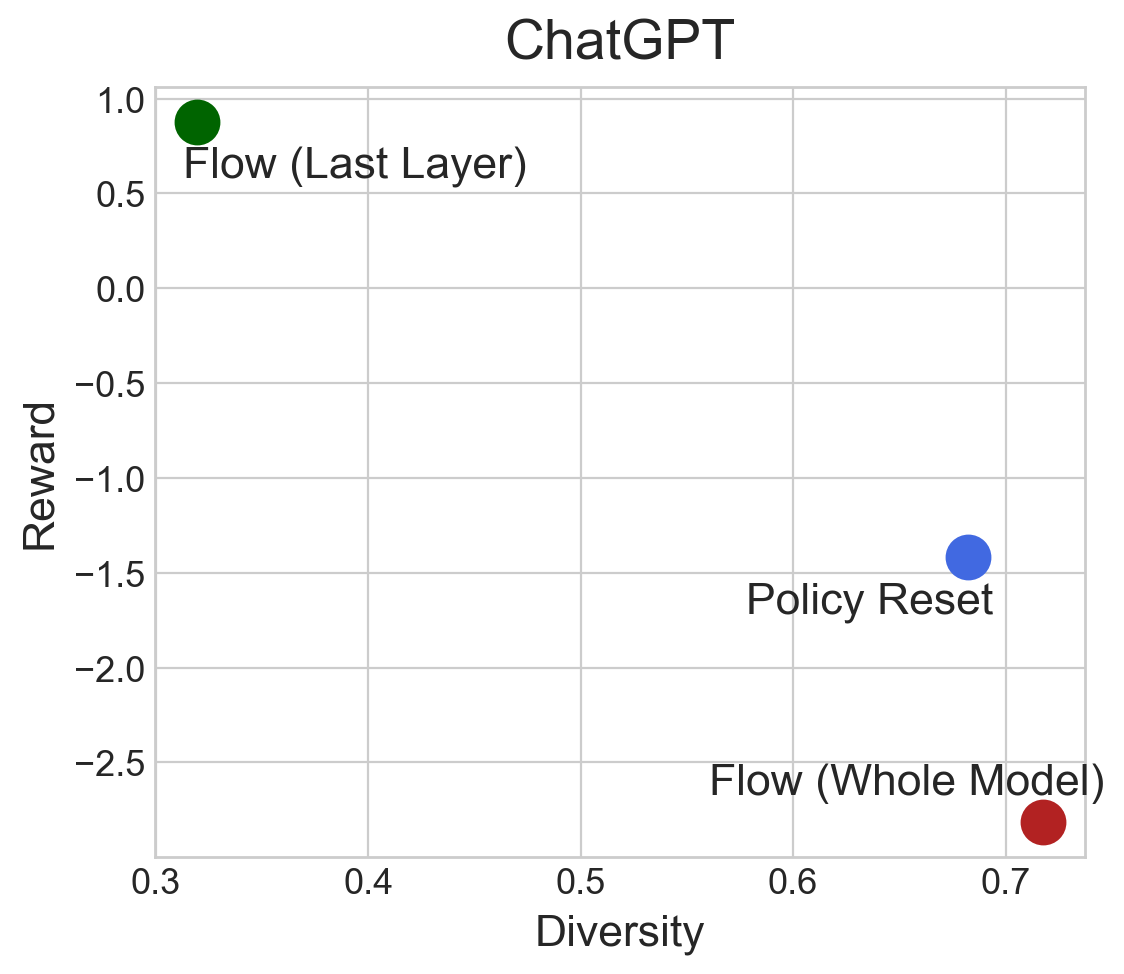}
        \subcaption{Analysis on reset strategies}
        \label{fig:ablation_reset}
    \end{subfigure}
    \caption{Extended ablation studies on various components of PAG.}
    \label{fig:ablation_app}
\end{minipage}
\end{figure*}
\section{Extended Main Results}\label{app:extend_main_results}
In this section, we provide additional discussion and analysis of our main experimental results.
\subsection{Main Results}
We summarize the main experiment results in \Cref{tab:main1} including comparisons with DPO-Diff \citep{wang2024discrete}, a recent relevant work. Note that as DPO-Diff tries to optimize negative prompts and the initial prompt is always the same, it is meaningless to compute diversity between generated prompts.
\subsection{Discussion}
As shown in the table, we observe that DPO-Diff shows modest improvements in terms of the reward compared to other baselines. We find that while DPO-Diff can effectively improve the CLIP scores by optimizing negative prompts, it shows limited capability in improving the aesthetic score.   

\section{Extended Ablation Studies}\label{app:extend_ablation}
In this section, we include additional ablation studies that complement our main text due to space limitations.

\subsection{Analysis on $\beta$}
First, we analyze the effect of inverse temperature $\beta$ in \cref{eq:reward}, which controls the balance between the task reward term $r(\mathbf{x}, \mathbf{y})$ and the reference LM likelihood term $p_{\text{ref}}(\mathbf{y}\vert\mathbf{x})$. As a default setting, we set $\beta=0.05$. 

To analyze the effect of $\beta$, we fine-tune the GFlowNet policy with different $\beta$ values: $\{0.01, 0.05, 0.1, 0.2\}$. As shown in the \Cref{fig:ablation_beta}, there are no significant differences in terms of the performance with different $\beta$ values while using too small $\beta$, which leads to the policy focus on a high-reward region, suffers from mode collapse similar to naive GFlowNet and exhibits poor performance. 

\begin{figure*}[t]
    \centering
    \includegraphics[width=0.9\textwidth]{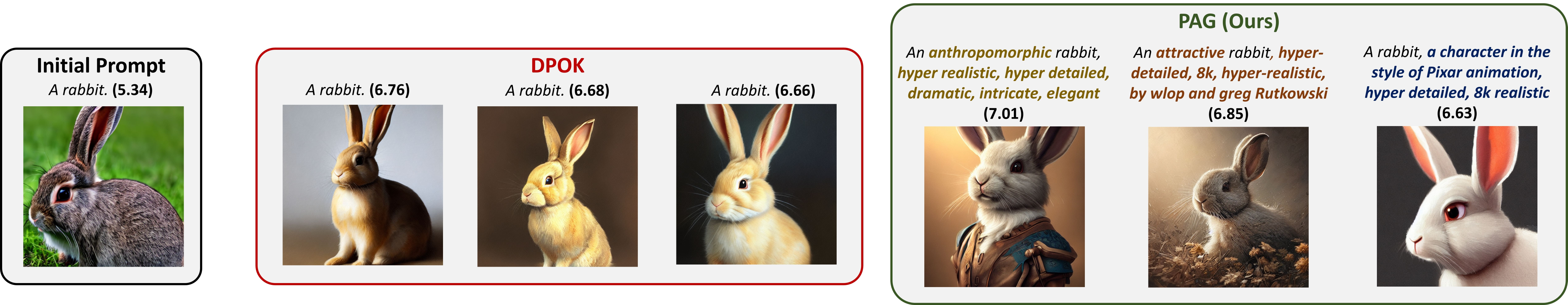}
    \caption{Comparison with DPOK and PAG. We report the aesthetic score of images in bold.}
    \label{fig:dpok_comparison}
\end{figure*}

\subsection{Analysis on $M$}
We also conduct experiments by varying the flow function reset period ($M$) to analyze the effect of flow reactivation. If we reset the flow function too frequently, it is hard to capture high-rewarding multi-modal distribution, while rarely applying reset leads to the mode collapse issue. As a default setting, we use $M=2000$, which means that we reactivate the flow function four times over the whole training procedure.  

To analyze the effect of $M$, we fine-tune the GFlowNet policy with different $M$ values: $\{1000, 2000, 4000\}$. As shown in the \Cref{fig:ablation_M}, we find that too frequent reactivation ($M=1000$) does not capture high-reward regions and suffers from a significant drop in performance. While there is no big difference between $M=2000$ and $M=4000$, we empirically find that set $M=2000$ achieves the best performance in terms of both reward and diversity. This empirical finding is also aligned with the other papers, which utilize a reset strategy: \citet{nikishin2022primacy} also reset the last layer of the neural networks four times over the course of training.

\subsection{Analysis on learning rate of flow function}
Based on the observation that most actor-critic RL methods use slightly higher learning rates for the critic than the actor \citep{schulman2017proximal}, we use a higher learning rate ($1\times10^{-4}$) for the flow function training than the learning rate of the policy ($1\times10^{-5}$). Using a higher learning rate is also crucial for quickly recovering from the flow reactivation.

To analyze the effect of learning rate ($\gamma$) for the flow function, we fine-tune the GFlowNet policy with different $\gamma$ values: $\{1\times10^{-3}, 1\times10^{-4}, 1\times10^{-5}\}$. As shown in the \Cref{fig:ablation_gamma}, we find that using the same learning rate for the policy and the flow function leads to poor performance as expected. While there is no big difference between $\gamma=1\times10^{-3}$ and $\gamma=1\times10^{-4}$, we empirically find that set $\gamma=1\times10^{-4}$ achieves the best performance in terms of both reward and diversity. 

\subsection{Analysis on flow reactivation scheme}
To prevent a significant drop in performance and unstable training, we employ a targeted reset strategy that resets only the last layer of the flow function. To analyze the effect of our strategy, we conduct experiments with two additional reset strategies: (1) reset the whole layer of the flow function and (2) reset the policy. For resetting the policy, we randomly reset $0.01\%$ of neurons in the policy parameters.

\Cref{fig:ablation_reset} shows the performance of various reset strategies. As depicted in the figure, resetting the whole layer of the flow function does not recover policy towards high-scoring regions. Resetting the policy parameters also exhibits poor performance, as the policy directly affects the acquisition of online samples.

\subsection{Analysis on Diversity of Images}
We also compute the diversity between the final images sampled from text-to-image diffusion models conditioned on prompts generated by different methods. To compute diversity, we compute the average pairwise distance between feature vectors extracted by the pre-trained InceptionV3 model \citep{szegedy2016rethinking}. As shown in \Cref{tab:div_images}, PAG exhibits high diversity on images compared to baselines.
\begin{table}[h]
\centering
\caption{Diversity evaluation on images.}
\vspace{-10pt}
\resizebox{\linewidth}{!}{
\begin{tabular}{lcc|cc}
\toprule
\multirow{2}{*}{\textbf{Method}} & \multicolumn{2}{c}{COCO} & \multicolumn{2}{c}{ChatGPT}  \\
\cmidrule{2-5}
& Reward & Diversity & Reward & Diversity \\
\midrule
SFT & 0.54 ± 0.07 & 0.20 ± 0.01 & 0.76 ± 0.03 & 0.19 ± 0.01 \\
Promptist & 0.65 ± 0.02 & 0.19 ± 0.01 & 0.76 ± 0.03 & 0.17 ± 0.01 \\
Rule-Based & 0.75 ± 0.01 & 0.19 ± 0.00 & 0.82 ± 0.03 & 0.18 ± 0.01 \\
GFlowNets & 0.73 ± 0.02 & 0.18 ± 0.01 & 0.63 ± 0.03 & 0.16 ± 0.01 \\
\midrule
PAG (Ours) & \textbf{0.88 ± 0.02} & \textbf{0.21 ± 0.01} & \textbf{0.88 ± 0.04} & \textbf{0.20 ± 0.01}\\
\bottomrule
\end{tabular}}
\vspace{-15pt}
\label{tab:div_images}
\end{table}

\section{Comparison with Directly Fine-tuning Diffusion Models}\label{app:comp_diffusion}
In this section, we explain in detail the comparison with directly fine-tuning diffusion models to generate images with desired properties.
\subsection{Experiment Setup}
We strictly follow the experiment setup of DDPO\footnote{https://github.com/jannerm/ddpo} and DPOK\footnote{https://github.com/google-research/google-research/tree/master/dpok} for fine-tuning diffusion models. We fine-tune diffusion models with aesthetic quality as a reward function and use prompts from a list of 45 common animals. 

\subsection{Comparison with DPOK}
We also compare our method with DPOK \citep{fan2024reinforcement}, another representative method for fine-tuning diffusion models with black-box reward functions. As shown in \Cref{fig:dpok_comparison}, generated images from DPOK converge to similar styles, whereas PAG generates diverse and high-quality images.

\section{Additional Visualizations}\label{app:more_visualization}
\begin{figure*}[t]
    \centering
    \includegraphics[width=0.9\textwidth]{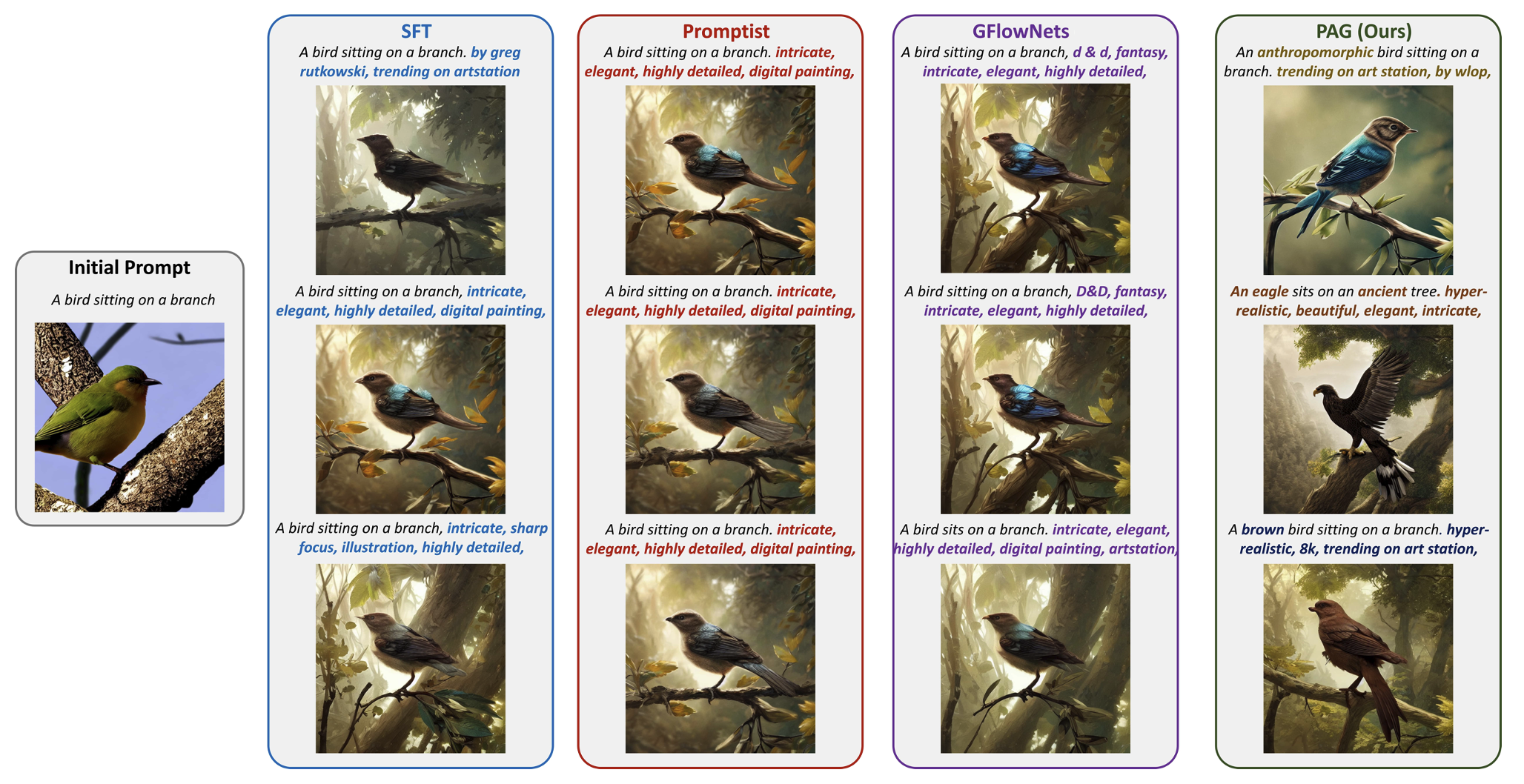}
    \caption{Additional images generated by optimized prompts using Stable Diffusion v1.4. We use the same seed to visualize the effect solely on prompt adaptation.}
    \label{fig:main_figure_type2}
\end{figure*}

\begin{figure*}[t]
    \centering
    \includegraphics[width=0.9\textwidth]{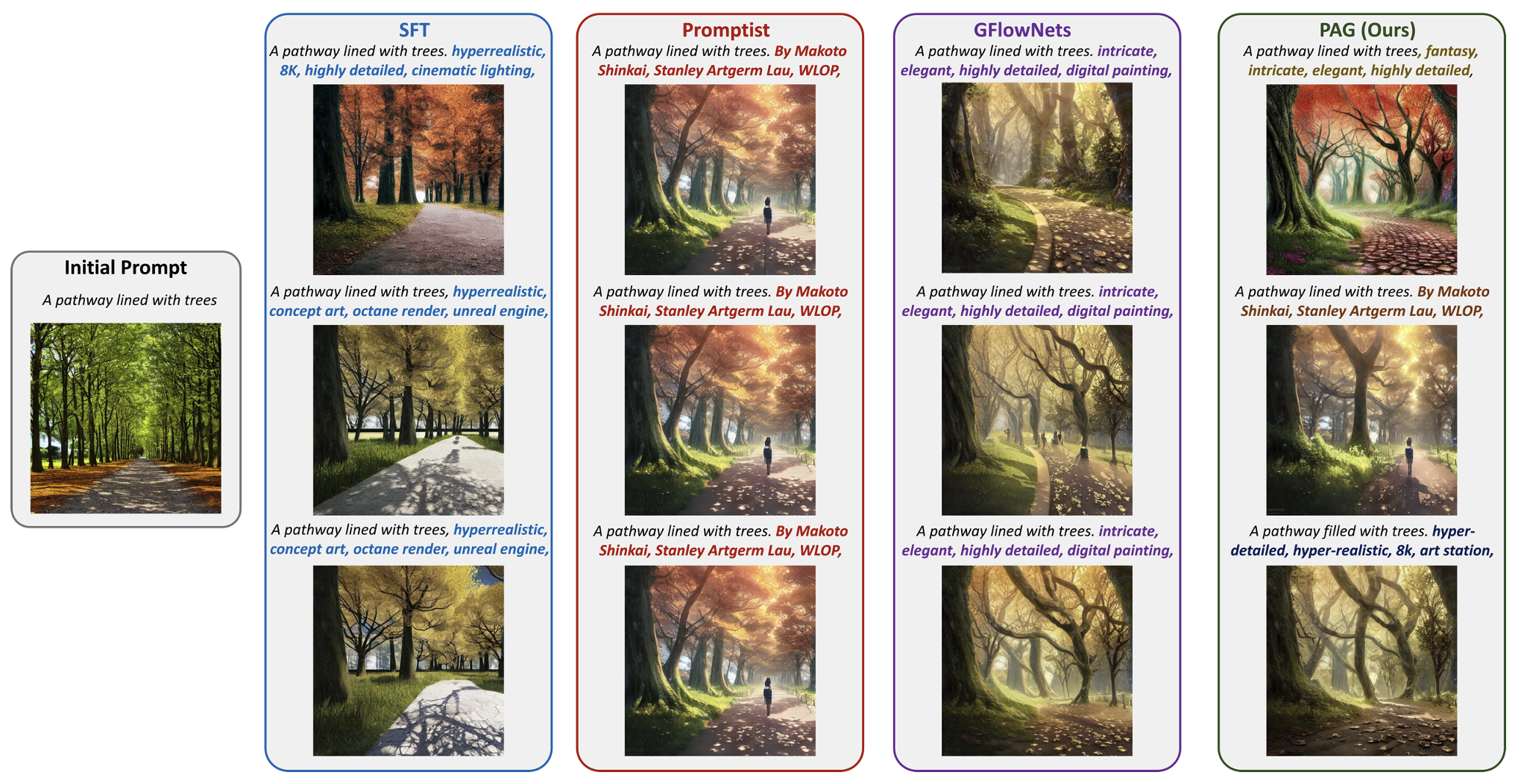}
    \caption{Additional images generated by optimized prompts using Stable Diffusion v1.4. We use the same seed to visualize the effect solely on prompt adaptation.}
    \label{fig:main_figure_type3}
\end{figure*}

\begin{figure*}[t]
    \centering
    \includegraphics[width=0.9\textwidth]{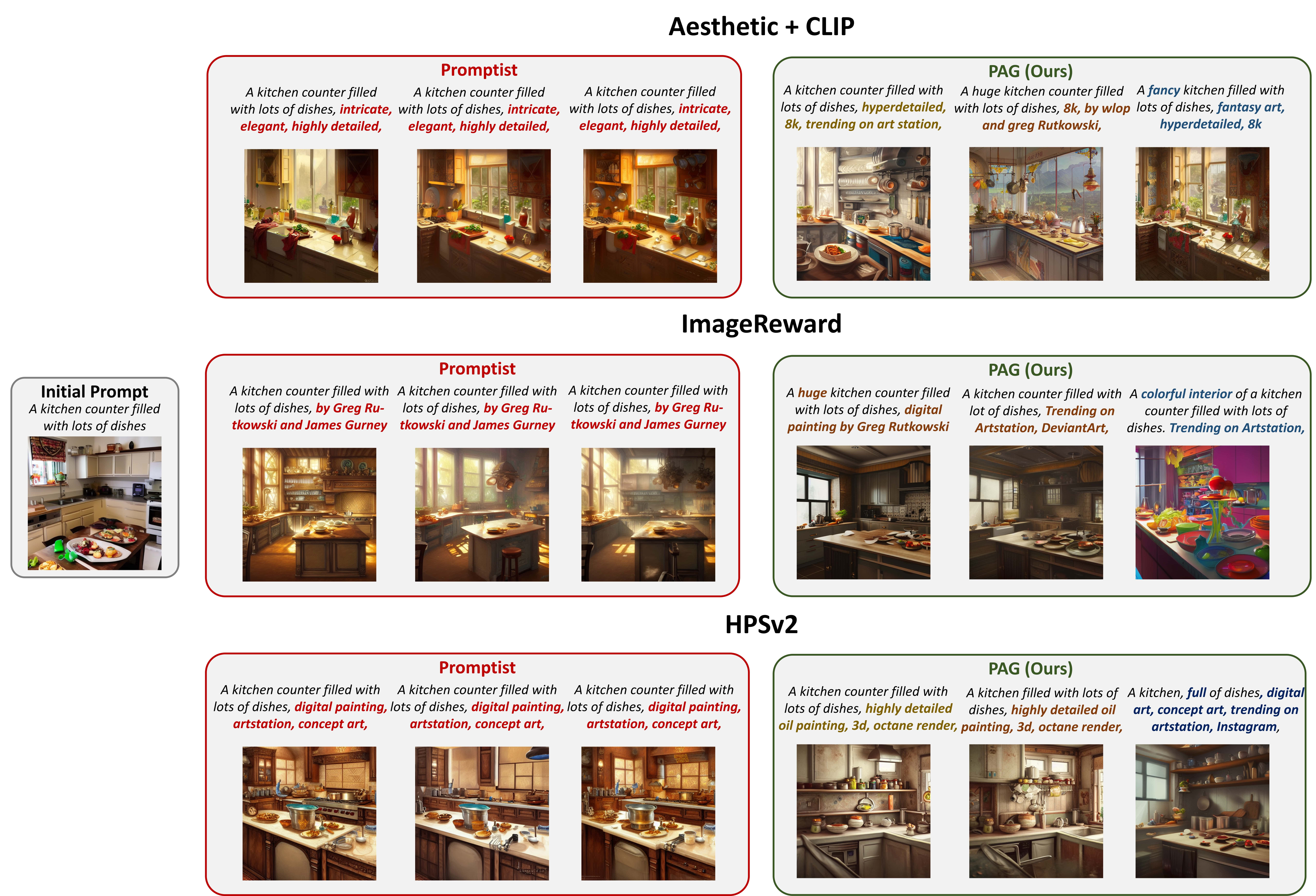}
    \caption{Images generated with prompts fine-tuned with different reward functions. We use the same seed to visualize the effect solely on prompt adaptation.}
    \label{fig:reward_fn}
\end{figure*}

\begin{figure*}[t]
    \centering
    \includegraphics[width=0.9\textwidth]{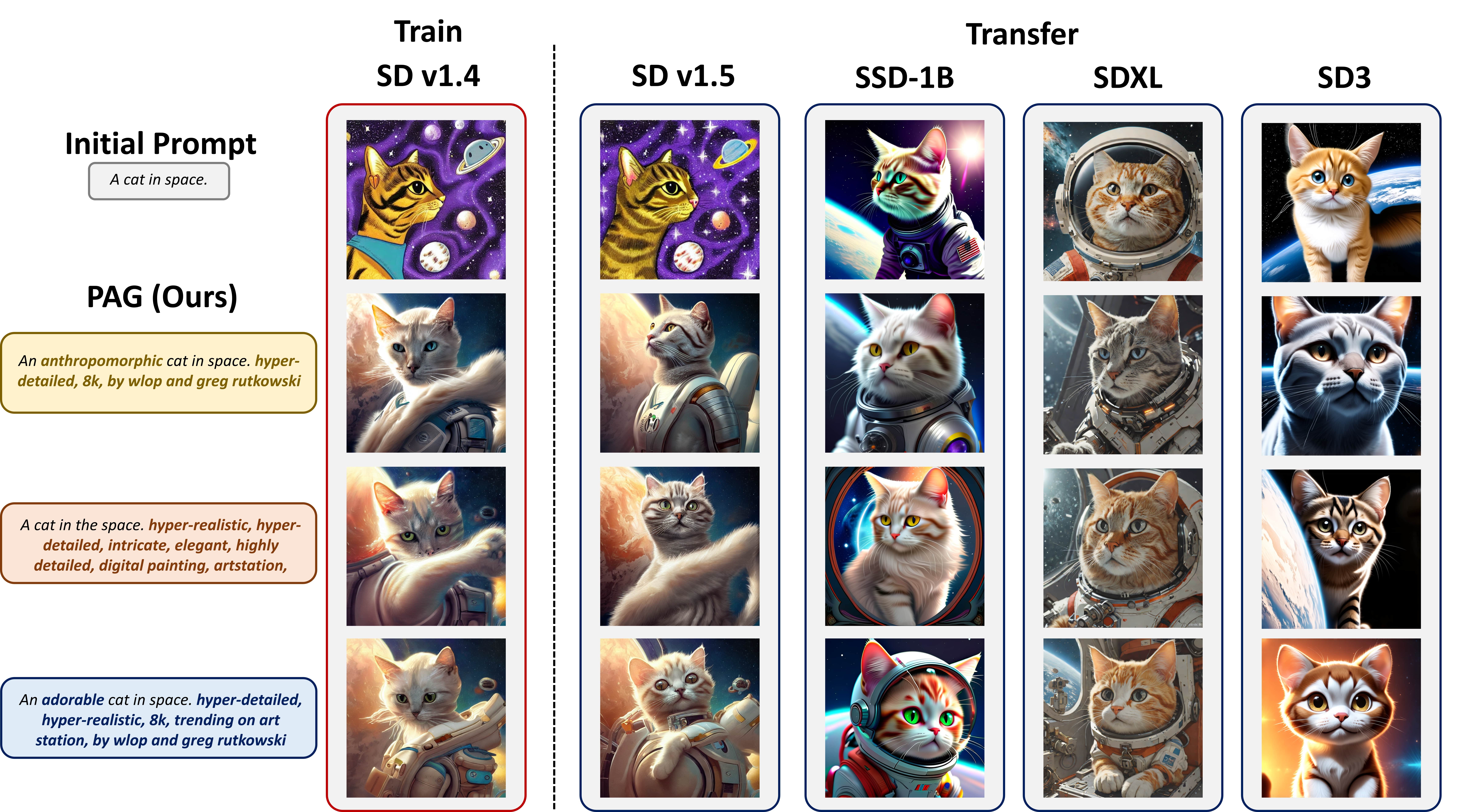}
    \caption{Images generated with different text-to-image diffusion models. We use the same seed to visualize the effect solely on prompt adaptation.}
    \label{fig:transfer}
\end{figure*}

In this section, we present additional visualizations to show the effectiveness of PAG for text-to-image generation as shown in Figures~\ref{fig:main_figure_type2}-\ref{fig:main_figure_type3} (besides Figure~\ref{fig:main_figure} in the main text). We also summarize the results for robustness across different reward functions and transferability to different text-to-image diffusion models as shown in Figure~\ref{fig:reward_fn}-\ref{fig:transfer}.

\end{document}